\documentclass{tlp}
\usepackage{times}
\usepackage{helvet}
\usepackage{courier}
\usepackage{graphicx}

\usepackage{latexsym,amssymb,url}
\usepackage{xspace,epsfig}

\def\beq{ 

\vspace*{-.5\baselineskip}\begin{equation}}
\def\eeq#1{\label{#1}\vspace{0ex}\end{equation}

\vspace*{-.25\baselineskip}
}

\def\bdm{

\vspace*{-.5\baselineskip}\begin{displaymath}}

\def\edm{\end{displaymath}

\vspace*{-.25\baselineskip}
}
\def\ba{\begin{array}}
\def\ea{\end{array}}

\def\bitem{\vspace{0ex}\begin{itemize}}
\def\eitem{\end{itemize}\vspace{0ex}}

\newtheorem{example}{Example}
\newtheorem{definition}{Definition}
\newtheorem{proposition}{Proposition}
\newtheorem{theorem}{Theorem}
\newtheorem{corollary}{Corollary}
\newtheorem{lemma}{Lemma}
\newcommand{\qed}{\hspace*{1em}\hbox{\proofbox}}

\newcommand{\OR}{\vee}
\newcommand{\AND}{\wedge}
\newcommand{\IMP}{\rightarrow}
\newcommand{\HT}{{HT}}
\newcommand{\QHT}{{QHT}}
\newcommand{\QEL}{{QEL}}
\newcommand{\sig}{{\cal L}}
\newcommand{\Ap}{A^+}
\newcommand{\An}{A^-}

\newcommand{\Lp}{L^+}
\newcommand{\Ln}{L^-}
\newcommand{\U}{{\cal U}}
\newcommand{\C}{{\cal C}}
\newcommand{\FS}{{\cal F}}
\newcommand{\PS}{{\cal P}}
\newcommand{\T}{\Gamma}
\newcommand{\F}{\phi}
\newcommand{\G}{\psi}
\newcommand{\p}{\Pi}
\newcommand{\gr}{\ensuremath{\mathit{grd}}}

\newcommand{\HU}{{\mathcal H}}
\newcommand{\HE}{{\mathcal E}}

\newcommand{\HB}{{\mathcal B}}

\newcommand{\BU}{\ensuremath{\HB_{\PS,\C_\U}}}
\newcommand{\sqhte}{\ensuremath{\mathbf{QHT^s_=}}}
\newcommand{\sqht}{\ensuremath{\mathbf{QHT^s}}}

\newcommand{\at}{{\mathcal A}}

\newcommand{\head}[1]{H(#1)}
\newcommand{\headp}[1]{H^+(#1)}
\newcommand{\headn}[1]{H^-(#1)}

\newcommand{\body}[1]{B(#1)}
\newcommand{\bodyp}[1]{B^+(#1)}
\newcommand{\bodyn}[1]{B^-(#1)}
\newcommand{\la}{\leftarrow}
\newcommand{\ra}{\rightarrow}

\newcommand{\naf}{\neg}

\newcommand{\Eh}[1]{E^{\Ap}_{\An}(#1)}
\newcommand{\Ehl}[1]{E^{\Lp}_{\Ln}(#1)}
\newcommand{\equivc}{\equiv_{c}}
\newcommand{\equiva}{\equiv_{a}}
\newcommand{\equivu}{\equiv_{u}}
\newcommand{\equivs}{\equiv_{s}}
\newcommand{\equive}{\equiv_{e}}
\newcommand{\equivh}{\ ^{\Ap}_{\An}\!\!\equiv\,}
\newcommand{\nequivh}{\ ^{\Ap}_{\An}\!\!\not\equiv}

\newcommand{\equivhl}{\ ^{\Lp}_{\Ln}\!\!\equiv\,}

\newcommand{\commadots}[0]{,\ldots ,}

\newcommand{\nop}[1]{}
\newcommand{\tuple}[1]{\ensuremath{\langle#1\rangle}}
\newif\ifmakebbl

\title[Equivalences in ASP by Countermodels in the Logic of Here-and-There]
{A General Framework for Equivalences in Answer-Set Programming by 
Countermodels \\
in the Logic of Here-and-There
\thanks{A preliminary version of this work appeared in the Proceedings of 
the 24\textsuperscript{th} International Conference on Logic Programming 
(ICLP), M.~Garcia de la Banda and E.~Pontelli (Eds.), LNCS 5366, pp.~99--113, 
Springer, 2008.}
}
\author[M. Fink]{
MICHAEL FINK \\
Institut f\"{u}r Informationssysteme, Technische Universit\"at Wien,\\
Favoritenstra{\ss}e 9-11, A-1040 Vienna, Austria. \\ 
\email{fink@kr.tuwien.ac.at}
}

\submitted{3\textsuperscript{rd} July 2009 }
\revised{16\textsuperscript{th} October 2009}
\accepted{8\textsuperscript{th} December 2009}

\begin{document}
\maketitle

\begin{abstract}
Different notions of equivalence, such as the prominent notions of strong 
and uniform equivalence, have been studied in Answer-Set Programming, 
mainly for the purpose of identifying programs that can serve as substitutes 
without altering the semantics, for instance in program optimization. 
Such semantic comparisons are usually characterized by various selections of 
models in the logic of Here-and-There (\HT). For uniform equivalence however, 
correct characterizations in terms of \HT-models can only be obtained for 
finite theories, respectively programs. In this article, we show that a 
selection of countermodels in \HT\ captures uniform equivalence also for 
infinite theories. 
This result is turned into coherent characterizations of the different notions 
of equivalence by countermodels, as well as by a mixture of \HT-models and 
countermodels (so-called equivalence interpretations).
Moreover, we generalize the so-called notion of relativized hyperequivalence 
for programs to propositional theories, and apply the same methodology in 
order to obtain a semantic characterization which is amenable to infinite 
settings. 
This allows for a lifting of the results to first-order theories under a 
very general semantics given in terms of a quantified version of \HT. 
We thus obtain a general framework for the study of various notions of 
equivalence for theories under answer-set semantics.
Moreover, we prove an expedient property that allows for 
a simplified treatment of extended signatures,
and provide further results for non-ground logic 
programs. In particular, uniform equivalence coincides under open and 
ordinary answer-set semantics, and for finite non-ground programs under 
these semantics, also the usual characterization of uniform equivalence in 
terms of maximal and total \HT-models of the grounding is correct, even for 
infinite domains, when corresponding ground programs are infinite. 

\medskip
\noindent
\textit{To appear in Theory and Practice of Logic Programming (TPLP).}

\medskip
\begin{keywords}
answer-set programming, uniform equivalence, relativized equivalence, 
knowledge representation.
\vspace*{-3ex}
\end{keywords}%
\end{abstract}

\section{Introduction} \label{sec:intro}

Answer-Set Programming (ASP) is a fundamental paradigm for nonmonotonic 
knowledge representation~\cite{bara-03} that encompasses logic programming 
under the answer-set semantics. It is distinguished by a purely declarative 
semantics and efficient 
solvers, such as, e.g., DLV~\cite{leon-etal-06}, Smodels~\cite{simo-etal-2002}, clasp~\cite{gebs-etal-07}, GnT~\cite{janh-niem-04}, and ASSAT~\cite{lin-zhao-04}.
Initially providing a semantics for rules with default negation in the body, 
the answer-set semantics (or stable-model semantics)~\cite{gelf-lifs-91}  
has been continually extended in terms of expressiveness and syntactic 
freedom. Starting with disjunctive rules, allowing for disjunctions in rule 
heads, negation in rule heads was considered and the development continued by 
allowing nested expressions, i.e., implication-free propositional formulas in 
the head and the body. Eventually, arbitrary propositional theories were given 
a non-classical minimal model semantics as their answer sets, which has 
recently been lifted to a general answer-set semantics for first-order 
theories~\cite{ferr-etal-07}.

In a different line of research, the restriction to Herbrand domains for 
programs with variables, i.e., non-ground programs, has been relaxed in order 
to cope with open domains~\cite{heym-etal-07}, which is desirable for certain 
applications, e.g., in conceptual modelling and Semantic Web reasoning. 
The resulting open answer-set semantics has 
been further generalized by dropping the unique names 
assumption~\cite{heym-etal-08} for application settings where it does not 
apply, for instance, when combining ontologies with 
nonmontonic rules~\cite{brui-etal-07}.

As for a logical characterization of the answer-set semantics, the logic of 
Here-and-There (\HT), a nonclassical logic extending intuitionistic logic, served 
as a basis.  Equilibrium Logic selects certain minimal \HT-models for 
characterizing the answer-set semantics for propositional theories and 
programs. It has recently been extended to Quantified Equilibrium Logic (\QEL) 
for first-order theories on the basis of a quantified version of Here-and-There 
(\QHT)~\cite{pear-valv-06,pear-valv-08}. 
Equilibrium Logic 
serves as a viable formalism for the study of semantic comparisons 
of theories and programs, like different notions of 
equivalence~\cite{EFTW05,lifs-etal-07,wolt-08,fabe-konc-06,fabe-etal-08,inou-saka-04}. 
The practical relevance of this research originates in program optimization tasks 
that rely on modifications that preserve certain 
properties~\cite{eite-etal-06,lin-chen-07,janh-etal-09,janh-08,saka-inou-09}.

In previous work~\cite{fink-08}, we complemented this line of research by 
solving an open problem concerning uniform equivalence of propositional theories 
and programs. 
Intuitively, two propositional logic programs are uniformly equivalent if they have 
the same answer sets under the addition of an arbitrary set of atoms to both 
programs.
Former characterizations of uniform equivalence, i.e., selections of \HT-models 
based on a maximality criterion~\cite{eite-etal-07a}, failed to 
capture uniform equivalence for infinite propositional programs---a problem 
that becomes relevant when turning to the non-ground setting, respectively 
first-order theories, where infinite domains, such as the natural numbers, 
are encountered in many application domains. In~\cite{fink-08}, this has been 
remedied resorting to countermodels in \HT.

In this article, we extend the former work beyond the basic notions of 
strong and uniform equivalence. So-called 
relativized notions thereof have been 
considered in order to capture more fine-grained semantical 
comparisons (see e.g.,~\cite{eite-etal-07a,pear-etal-07}). 
Intuitively, these notions restrict the alphabet to be considered for 
potential additions, i.e., programs or sets of facts, respectively. A further 
refinement distinguishes the alphabet for atoms allowed in rule heads 
of an addition from the alphabet for atoms allowed in rule 
bodies~\cite{wolt-08}.
The various notions of equivalence that can be formalized this way have 
recently been called \emph{relativized hyperequivalence}~\cite{trus-wolt-08b,trus-wolt-08c}.

Similarly as for uniform equivalence, semantic characterizations of 
relativized hyperequivalence have been obtained by means of a maximality 
criterion so far, and only for finite propositional settings. We address this issue
and apply the same methods as for uniform equivalence in order to obtain alternative 
characterizations. They can be stated without any finiteness restrictions and easily 
lift to first-order settings over infinite domains.

The new contributions compared to~\cite{fink-08} can be summarized as follows:
\begin{itemize}
\item We provide full proofs for the characterizations of uniform equivalence, 
but also classical equivalence, answer-set equivalence, and strong equivalence, 
in terms of countermodels in \HT, respectively in terms of 
\emph{equivalence interpretations}, developed in~\cite{fink-08}.

\item We extend these ideas to relativized settings of equivalence and generalize 
the notion of relativized hyperequivalence to propositional 
theories. Abstracting from the notions of rule head and rule body, we obtain 
respective notions of relativization for theories. 
We provide novel semantical characterizations in terms of equivalence 
interpretations for this generalized setting, again without any finiteness 
restrictions.

\item We lift these results to first-order theories by means of \QHT, 
essentially introducing, besides uniform equivalence, relativized hyperequivalence 
for first-order theories under the most 
general form of answer-set semantics currently considered. 

\item We correct an informal claim that has been made in connection with a 
property which allows for a simplified treatment of extended signatures and holds 
for \QHT\ countermodels.
Based on an erroneous example (Example~5 in~\cite{fink-08}), it was claimed that 
this property does not hold for \QHT-models, which is not the case.

\item 
Eventually, we reconsider logic programs and prove, 
using
the established characterization, that uniform equivalence coincides for 
open and ordinary answer-set semantics, 
as well as other results which 
have been stated without proof in~\cite{fink-08}.
\end{itemize}

Our results provide an elegant, uniform model-theoretic framework for the
characterization of the different notions of equivalence considered in ASP.
They generalize to first-order theories without 
finiteness restrictions, and are relevant for practical 
ASP systems that handle finite non-ground programs over infinite domains.

In particular relativized notions of equivalence are  
relevant in practice. For instance, program composition from modular parts 
is an issue of increasing interest in ASP~\cite{dao-etal-09a,janh-etal-09}.
It usually hinges on semantic properties specified for an interface 
(input/output for `calling' or connecting modules), i.e., properties that 
require compliance on a subset of the underlying language. Our results might 
be exploited to provide correctness guarantees for specific compositions.

Another benefit comes with the generalization to first-order theories. It 
facilitates and simplifies the study of combinations of ASP with 
other formalisms, or means for external data access, in a unifying formalism. 
Especially the combination of nonmonotonic rules with description logics is a 
highly relevant instance of such a combination. Our results can initiate or 
reduce difficulties in the study of modularity and optimization for such 
combined settings.
(cf.~\cite{fink-pear-09} for preliminary work in this direction).

For the sake of presentation, the technical content is split into two parts, 
discussing the propositional case first, 
and addressing first-order theories and nonground programs in 
a second part. 
In particular, the organization is as follows: Section~\ref{sec:prelims} 
introduces essential preliminaries for the treatment of the propositional 
case. In Section~\ref{sec:prop}, we develop a characterization of uniform 
equivalence by means of countermodels in \HT, and proceed with an alternative 
characterization in terms of equivalence interpretations, before we turn to 
generalizing and characterizing relativized hyperequivalence for 
propositional theories. After some introductory background on quantified \HT, 
Section~\ref{sec:nonground} deals with generalizations of previous results 
to first-order theories under generalized answer-set semantics. In 
Section~\ref{sec:lp}, we apply our characterization of uniform equivalence 
to logic programs under various extended semantics in comparison with the 
traditional semantics over Herbrand domains, before we draw some conclusions
in Section~\ref{sec:conclusion}.

\section{Preliminaries} \label{sec:prelims}

We start with the propositional setting and briefly summarize the necessary 
background. Corresponding first-order formalisms will be introduced when discussing 
first-order theories, respectively non-ground logic programs.

\subsection{Propositional Here-and-There}
In the propositional case we consider formulas of a propositional signature
$\sig$, i.e., a set of propositional variables, 
and the connectives $\AND$, $\OR$, $\IMP$, and $\bot$
for conjunction, disjunction, implication, and falsity, respectively.
Furthermore we make use of the following abbreviations: $\F\equiv \G$ for 
$(\F\IMP \G) \AND (\G\IMP \F)$; $\neg \F$ for $\F\IMP\bot$; and $\top$ for 
$\bot\IMP\bot$.
A formula is said to be {\em factual}\footnote{When uniform equivalence of theories 
is considered, then factual theories can be considered instead of facts---hence 
the terminology---see also the discussion at the end of this section.} 
if it is built using  
$\AND$, $\OR$, $\bot$, and $\neg$ (i.e., implications of the form $\F\IMP\bot$), 
only.
A theory $\T$ is factual if every formula of $\T$ has this property.

The logic of here-and-there is an intermediate logic between intuitionistic logic 
and classical logic. Like intuitionistic logic it can be semantically characterized 
by Kripke models, in particular using just two worlds, namely ``{\em here\/}'' and 
``{\em there\/}'' 
(assuming that the {\em here\/} world is ordered before the {\em there\/} world). 
Accordingly, 
interpretations (\HT-interpretations) 
are pairs $(X,Y)$ of sets of atoms from $\sig$, such that $X\subseteq Y$. 
An \HT-interpretation is {\em total} if $X=Y$. The intuition is that atoms in $X$ 
(the {\em here\/} part) are considered to be true, atoms not in $Y$ 
(the {\em there\/} part)
are considered to be false, while the remaining atoms (from $Y\setminus X$) are 
undefined. 

We 
denote classical satisfaction of a formula $\F$ by an interpretation $X$, 
i.e., a set of atoms, as $X\models \F$, whereas satisfaction in the logic of 
here-and-there (an \HT-model), symbolically  $(X,Y)\models \F$, is defined recursively:
\begin{enumerate}
\item $(X,Y)\models a$ if $a\in X$, for any atom $a$,
\item $(X,Y)\not\models \bot$,
\item $(X,Y)\models \F\AND \G$ if $(X,Y)\models \F$ and $(X,Y)\models \G$,
\item $(X,Y)\models \F\OR \G$ if $(X,Y)\models \F$ or $(X,Y)\models \G$,
\item $(X,Y)\models \F\IMP \G$ if ($i$) $(X,Y)\not\models \F$ or $(X,Y)\models \G$, 
and ($ii$) $Y\models \F\IMP \G$\footnote{That is, $Y$ satisfies $\F\IMP \G$ classically.}.
\end{enumerate}

An \HT-interpretation $(X,Y)$ satisfies a theory $\T$, iff it satisfies all formulas 
$\F\in \T$. For an axiomatic proof system see, 
e.g.,~\cite{lifs-etal-07}.

A total \HT-interpretation $(Y,Y)$ is called an \emph{equilibrium model} of a theory 
$\T$, iff $(Y,Y)\models \T$ and for all \HT-interpretations $(X,Y)$, such that 
$X\subset Y$, it holds that $(X,Y)\not\models\T$. 
An interpretation $Y$ is an \emph{answer set} of $\T$ iff
$(Y,Y)$ is an equilibrium model of $\T$.

We will 
make use of the following simple properties: 
if $(X,Y)\models \T$ then $(Y,Y)\models \T$; 
and $(X,Y)\models \neg \F$ iff $Y\models \neg \F$;  
as well as of the following lemma.

\begin{lemma}[Lemma~5 in~\cite{Pearce04}]
\label{lemm:pearce} 
If $\F$ is a factual propositional formula, 
$(X,Y)\models \F$, and 
$X\subseteq X'\subseteq Y$, then $(X',Y)\models \F$.
\end{lemma}

\subsection{Propositional Logic Programming}

A (\emph{disjunctive}) 
rule $r$ is of the form 
\begin{equation}\label{form:rule}
\begin{array}{l}
a_1\vee\cdots\vee a_k \vee \naf a_{k+1}\vee\cdots\vee \naf a_l \ \la
\ b_1,\ldots, b_m,\naf b_{m+1},\ldots, \naf b_n,
\end{array}
\end{equation}

\noindent 
where $a_1,\ldots ,a_l,b_1,\ldots ,b_n$ are
atoms of a propositional signature $\sig$, such that
$l\geq k\geq 0$, $n\geq m\geq 0$, and $l+n > 0$.
We refer to ``$\naf$'' as {\em default negation}.
The \emph{head} of $r$ is the set
$\head{r} = \{a_1\commadots a_k, \naf a_{k+1}\commadots \naf a_l\}$, and
the \emph{body} of $r$ is denoted by 
$\body{r}=\{b_1,\ldots, b_m,\, \naf b_{m+1},\ldots,\, \naf b_n\}$.
Furthermore, we define the sets  
$\headp{r}=$ $\{a_{1}\commadots a_{k}\}$, 
$\headn{r}=$ $\{a_{k+1}\commadots a_l\}$, 
$\bodyp{r}=$ $\{b_{1}\commadots b_{m}\}$, and 
$\bodyn{r}=$ $\{b_{m+1}\commadots b_n\}$.
A \emph{program} $\p$ (over $\sig$) is a set of rules (over $\sig$). 

An interpretation $I$, i.e., a set of atoms, satisfies a rule $r$, symbolically 
$I\models r$, iff 
$I\cap\headp{r}\neq\emptyset$ or $\headn{r}\not\subseteq I$,  
if $\bodyp{r}\subseteq I$ and $\bodyn{r}\cap I=\emptyset$. 
Adapted from~\cite{gelf-lifs-91}, the \emph{reduct} 
of a program $\p$ with respect to 
an interpretation $I$, symbolically $\p^I$, 
is given by the set of rules 
$$a_1\vee\cdots\vee a_k \ \la\ b_1,\ldots, b_m, $$
obtained from rules in $\p$,  
such that $\headn{r}\subseteq I$ and $\bodyn{r}\cap I=\emptyset$.

An interpretation $I$ is called 
an \emph{answer set}
of $\p$ iff  
$I\models \p^I$ 
and it is subset minimal among the 
interpretations of $\sig$ 
with this property.

\subsection{Notions of Equivalence}
For any two theories, respectively programs, 
and a potential extension by $\T$, we consider the following notions of equivalence 
which have been shown to be the only forms of equivalence obtained by 
varying the logical form of extensions in the propositional case 
in~\cite{Pearce04}.

\begin{definition}
Two 
theories  $\T_1, \T_2$ over $\sig$ are called
\begin{itemize}
\item \emph{classically equivalent}, 
$\T_1\equivc\T_2$, 
if and only if they have the same classical models;
\item \emph{answer-set equivalent}, 
$\T_1\equiva\T_2$,  
if and only if they have the same answer sets, i.e., equilibrium models;
\item \emph{strongly equivalent}, 
$\T_1\equivs\T_2$, 
if and only if, for any theory $\T$ over $\sig'\supseteq \sig$, 
$\T_1\cup \T$ and  $\T_2\cup \T$ are answer-set equivalent;
\item \emph{uniformly equivalent}, 
$\T_1\equivu\T_2$,  
if and only if, for any factual theory $\T$ over $\sig'\supseteq \sig$, 
$\T_1\cup \T$ and  $\T_2\cup \T$ are answer-set 
equivalent.
\end{itemize}
\end{definition}

Emanating from a logic programming setting, uniform equivalence is usually 
understood wrt.~sets of \emph{facts} (i.e., atoms). 
Obviously, uniform equivalence wrt.~factual theories implies uniform 
equivalence wrt.~sets of facts. The converse direction has been shown as well 
for general propositional theories (cf.~Theorem~2 in~\cite{Pearce04}).
Therefore, in general there is no difference whether uniform equivalence is 
considered wrt.~sets of facts or factual theories. The latter may be 
regarded as facts, i.e., rules with an empty body, of so-called nested 
logic program rules. One might also consider sets of disjunctions of atomic 
formulas and their negations (i.e., clauses), accounting for facts 
according to the definition of program rules in this article. Note that 
clauses constitute factual formulas and the classical transformation of 
clauses into implications is not valid under answer set semantics 
(respectively in \HT). 

\section{Equivalence of Propositional Theories by \HT-Countermodels} \label{sec:prop}

Uniform equivalence is usually characterized by so-called UE-models, i.e., 
total and maximal non-total \HT-models, which fail to capture uniform 
equivalence for infinite propositional theories. 

\begin{example}[\cite{eite-etal-07a}]
\label{ex:inf}{\rm
Let $\T_1$ and $\T_2$ over 
$\sig=\{a_i \mid i\geq 1\}$ be the following propositional theories 
\vspace*{-1ex}
\begin{eqnarray*}
\T_1 = \{ a_i \mid i \geq 1 \},  & \textrm{ and } &  
\T_2 = \{ \neg a_i \ra a_i,\ a_{i+1} \ra a_i
\mid i \geq 1 \}. 
\end{eqnarray*}
Both, $\T_1$ and $\T_2$, have the single 
total \HT-model $(\sig,\sig)$. 
Furthermore, $\T_1$ has no non-total \HT-model $(X,\sig)$, i.e,  
such that $X\subset \sig$, while $\T_2$ has the non-total \HT-models $(X_i,\sig)$, 
where $X_i = \{a_1,\ldots,a_i\}$ for $i \geq 0$.
Both theories have the same total and maximal non-total (namely none)  
\HT-models. But they are not uniformly equivalent as witnessed by the fact that
$(\sig,\sig)$ is an equilibrium model of $\T_1$ but not of  $\T_2$.
\qed}\end{example}

The reason for this failure 
is the inability of the concept of maximality 
to capture differences exhibited by an infinite number of \HT-models.

\subsection{\HT-Countermodels} \label{sec:counter}

The above problem can be avoided by taking \HT-countermodels that 
satisfy a closure condition instead of the maximality criterion.

\begin{definition}
An \HT-interpretation $(X,Y)$ is an \emph{\HT-countermodel} of a theory $\T$ if 
$(X,Y)\not\models\T$. The set of \HT-countermodels of a theory $\T$ is denoted 
by $C_s(\T)$.
\end{definition}

Intuitively, an \HT-interpretation fails to be an \HT-model of a theory $\T$ 
when the theory is not 
satisfied at one of the worlds ({\em here\/} or {\em there\/}). Note that 
satisfaction at the {\em there\/}
world amounts to classical satisfaction of the theory by $Y$. 
A simple consequence is that if $Y\not\models\T$, then $(X,Y)$ is an \HT\ 
countermodel of $\T$ for any $X\subseteq Y$. At the {\em here\/} world, 
classical satisfaction is a sufficient condition but not necessary. For logic 
programs, satisfaction at the {\em here\/} world is precisely captured by the 
reduct of the program $\p$ wrt.~the interpretation at the {\em there\/} world, i.e., 
if $X\models \p^Y$.

\begin{definition} A total \HT-interpretation $(Y,Y)$ is 
\emph{total-closed} in a set $S$ of \HT-interpretations 
if $(X,Y)\in S$ for every $X\subseteq Y$.
We say that an \HT-interpretation $(X,Y)$ is
\begin{itemize}
\item \emph{closed} in a set $S$ of \HT-interpretations 
if $(X',Y)\in S$ for every $X\subseteq X'\subseteq Y$.
\item 
\emph{there-closed} in a set $S$ of \HT-interpretations 
if $(Y,Y)\not\in S$ and $(X',Y)\in S$ for every $X\subseteq X'\subset Y$.
\end{itemize}
\end{definition}

A set  $S$ of \HT-interpretations is total-closed, 
if every total \HT-interpretation $(Y,Y)\in S$ is total-closed in $S$. 
By the remarks on the satisfaction at the {\em there\/} world above, it is obvious 
that every total \HT-countermodel of a theory is also total-closed in $C_s(\T)$. 
Consequently, $C_s(\T)$ is a total-closed set for any theory $\T$. 
By the same argument, if $(X,Y)$ is an \HT-countermodel such that $X\subset Y$ 
and $Y\not\models\T$, then $(X,Y)$ is closed in $C_s(\T)$. The more relevant 
cases concerning the characterization of equivalence are \HT-countermodels $(X,Y)$ 
such that $Y\models\T$.

\begin{example}{\rm 
Consider the theory $\T_1$ in Example~\ref{ex:inf} and a non-total \HT-interpretation 
$(X,\sig)$. Since $(X,\sig)$ is non-total, $X\subset\sig$ holds, and therefore 
$(X,\sig)\not\models a_i$, for some $a_i\in\sig$. Thus, we have identified a \HT-countermodel
of $\T_1$. Moreover the same argument holds for any non-total \HT-interpretation of the from 
$(X',\sig)$ (in particular such that $X\subseteq X'\subset Y$). Therefore, $(X,\sig)$ is 
there-closed in $C_s(\T_1)$.
\qed}\end{example}

The intuition that, essentially, there-closed countermodels can be used instead of  
maximal non-total \HT-models for characterizing uniform equivalence draws from the 
following observation. If $(X,Y)$ is a maximal non-total \HT-model, then every 
$(X',Y)$, such that $X\subset X'\subset Y$, is a there-closed \HT-countermodel. 
However, there-closed \HT-countermodels are not sensitive to the problems that 
infinite chains cause for maximality.

Given a theory $\T$, let $C_u(\T)$ denote the set of there-closed 
\HT-interpretations in $C_s(\T)$.

\begin{theorem}\label{theo:counter}
Two 
propositional theories $\T_1$, $\T_2$ are uniformly equivalent iff 
they have the same sets of there-closed \HT-countermodels, 
in symbols $\T_1\equivu\T_2$ iff $C_u(\T_1)=C_u(\T_2)$.
\end{theorem}

\begin{proof}
For the only-if direction, assume that two theories, $\T_1$ and $\T_2$, are 
uniformly equivalent. Then they are classically equivalent, i.e., they 
coincide on total \HT-models, and therefore also on total \HT-countermodels. 
Since a total \HT-interpretation $(Y,Y)$ is there-closed in $C_s(\T)$ if 
$(Y,Y)\not\in C_s(\T)$, i.e., if $(Y,Y)$ is an \HT-model of $\T$, this proves 
that $\T_1$ and $\T_2$ coincide on total \HT-interpretations that are 
there-closed in $C_s(\T_1)$, respectively in $C_s(\T_2)$.

To prove our claim, it remains to show that $\T_1$ and $\T_2$ coincide on 
non-total there-closed \HT-countermodels $(X,Y)$, i.e., such that $(Y,Y)$ is 
an \HT-model of both theories. Consider such a there-closed \HT-countermodel of $\T_1$. Then, $(Y,Y)$ 
is a total \HT-model of $\T_1\cup X$
and no $X'\subset Y$ exists such that $(X',Y)\models \T_1\cup X$, either because it 
is an \HT-countermodel of $\T_1$ (in case $X\subseteq X'\subset Y$) or of $X$ 
(in case $X'\subset X$). Thus, $Y$ is an answer set of $\T_1\cup X$ and, by 
hypothesis since $X$ is factual, it is also an answer set of $\T_2\cup X$. 
The latter implies for all $X\subseteq X'\subset Y$ that 
$(X',Y)\not\models \T_2\cup X$. All these \HT-interpretations are \HT-models of 
$X$. Therefore we conclude that they all are \HT-countermodels of $\T_2$ and 
hence $(X,Y)$ is a there-closed \HT-countermodel of $\T_2$. Again 
by symmetric arguments, we establish the same for any  there-closed \HT\ 
countermodel $(X,Y)$ of $\T_2$ such that $(Y,Y)$ is a common total \HT-model. 
This proves that $\T_1$ and $\T_2$ have the same sets of there-closed \HT\ 
countermodels.

For the if direction, assume that two theories, $\T_1$ and $\T_2$, have the same 
sets of there-closed \HT-countermodels. This implies that they have the same 
total \HT-models (since these are there-closed). 
Consider any factual theory $\T'$ such that $Y$ is 
an answer set of $\T_1\cup \T'$. We show that $Y$ is an answer set of 
$\T_2\cup \T'$ as well. Clearly, 
$(Y,Y)\models \T_1\cup \T'$ implies $(Y,Y)\models \T'$ and therefore 
$(Y,Y)\models \T_2\cup \T'$. Consider any $X\subset Y$. Since $Y$ is an answer set 
of $\T_1\cup \T'$, it holds that $(X,Y)\not\models \T_1\cup \T'$. We show that 
$(X,Y)\not\models \T_2\cup \T'$. If $(X,Y)\not\models \T'$ this is trivial, 
and in particular the case if $(X,Y)\models \T_1$. So let us consider the 
case where $(X,Y)\not\models \T_1$ and $(X,Y)\models \T'$. By 
Lemma~\ref{lemm:pearce} we conclude from the latter that, for any  
$X\subseteq X'\subset Y$, 
$(X',Y)\models \T'$. Therefore, $(X',Y)\not\models \T_1$, as well. This implies that 
$(X,Y)$ is a there-closed \HT-countermodel of $\T_1$. By hypothesis, $(X,Y)$ is a 
there-closed \HT-countermodel of $\T_2$, i.e., $(X,Y)\not\models \T_2$. 
Consequently, $(X,Y)\not\models \T_2\cup \T'$. Since this argument applies to any 
$X\subset Y$, $(Y,Y)$ is an equilibrium model of  $\T_2\cup \T'$, i.e., $Y$ is an 
answer set of $\T_2\cup \T'$. The 
argument with $\T_1$ and $\T_2$ interchanged, 
proves that $Y$ is an answer set of $\T_1\cup \T'$ if it is an answer set of 
$\T_2\cup \T'$. Therefore, the answer sets of  $\T_1\cup \T'$ and $\T_2\cup \T'$ 
coincide for any factual 
$\T'$, i.e., $\T_1$ and $\T_2$ are uniformly 
equivalent.
\end{proof}

\begin{example}\label{ex:infctd1}{\rm
Reconsider the theories in Example~\ref{ex:inf}. Every non-total \HT-interpretation 
$(X_i,\sig)$ is an \HT-countermodel of $\T_1$, and thus, each of them is 
there-closed in $C_s(\T_1)$. On the other hand, none of these 
\HT-interpretations is an \HT-countermodel of $\T_2$. Therefore, $\T_1$ and 
$\T_2$ are not uniformly equivalent.
\qed}\end{example}

Countermodels have the drawback however, that they cannot be characterized 
directly in \HT\ 
itself, 
i.e., as the \HT-models of a `dual' theory. 
The usage of 
``dual'' here 
is 
non-standard compared to its application to particular calculi or 
consequence relations, but it likewise conveys the idea of a dual concept.
In this sense \HT\ 
therefore is non-dual:

\begin{proposition}\label{prop:nondual}
Given a theory $\T$, in general there is no theory $\T'$ such that
$(X,Y)$ is an \HT-countermodel of $\T$ iff it is a \HT-model of $\T'$, for any 
\HT-interpretation $(X,Y)$. 
\end{proposition}

\begin{proof}
As observed in~\cite{caba-ferr-07}, any theory has a 
total-closed set of countermodels. Consider the theory $\T=\{a\}$ and suppose 
there exists a theory $\T'$, such that $(X,Y)$ is an \HT-countermodel of $\T$ 
iff it is an \HT-model of $\T'$. Then, vice versa, $(X,Y)$ is an \HT-countermodel 
of  $\T'$ iff it is an \HT-model of $\T$. Since for $Y=\{a\}$, $(Y,Y)$ is an 
\HT-model of $\T$, we conclude that $(Y,Y)$ is an \HT-countermodel of $\T'$. 
Because any theory has a total-closed set of countermodels, it follows that 
$(\emptyset,Y)$ is an \HT-countermodel of $\T'$, hence, an \HT-model of $\T$. 
Contradiction.
\end{proof}

\subsection{Characterizing Equivalence by means of Equivalence Interpretations} \label{sec:equiv}

The characterization of countermodels by a theory in \HT\ essentially fails 
due to total \HT-countermodels. However, total \HT-countermodels of a theory 
are not necessary for characterizing equivalence, in the sense that they can 
be replaced by total \HT-models of the theory for this purpose. 

\begin{definition}
An \HT-countermodel $(X,Y)$ of a theory $\T$ is called a \emph{here-countermodel} 
of $\T$ if $Y\models \T$.
\end{definition}

\begin{definition}
An \HT-interpretation is an \emph{equivalence interpretation} of a theory $\T$ if it 
is a total \HT-model of $\T$ or a here-countermodel of $\T$. The set of 
equivalence interpretations of a theory $\T$ is denoted by $E_s(\T)$.
\end{definition}

\begin{theorem}\label{theo:equimod}
Two theories $\T_1$ and $\T_2$ coincide on their \HT-countermodels iff they 
have the same equivalence interpretations,
symbolically $C_s(\T_1)=C_s(\T_2)$ iff $E_s(\T_1)=E_s(\T_2)$.
\end{theorem}

\begin{proof}
For the only-if direction, assume that two theories, $\T_1$ and $\T_2$, have the 
same sets of \HT-countermodels. This implies that they have the same 
here-countermodels. Furthermore, since the total \HT-countermodels are equal, 
they also coincide on total \HT-models. Consequently, $\T_1$ and $\T_2$ have the 
same equivalence interpretations.

For the if direction, assume that two theories, $\T_1$ and $\T_2$, coincide on their 
equivalence interpretations. Then they have the same total \HT-models and hence the same 
total \HT-countermodels. Since total \HT-countermodels of every theory are 
total-closed in the set of \HT-countermodels, the sets of \HT-countermodels 
coincide on all \HT-interpretations $(X,Y)$ such that $(Y,Y)$ is a (total) \HT\ 
countermodel. All remaining \HT-countermodels are here-countermodels and 
therefore coincide by hypothesis and the definition of equivalence interpretations. 
This proves the claim.
\end{proof}

As a consequence of this result, and the usual relationships on \HT-models, we 
can characterize equivalences of propositional theories 
also by selections of equivalence interpretations, i.e., a mixture of non-total 
here-countermodels and total \HT-models, such that the characterizations, 
in particular for uniform equivalence, are also correct for infinite theories. 

\begin{definition}\label{def:eqsets}
Given a theory $\T$, we denote by  
\begin{itemize}
\item 
$C_c(\T)$, respectively $E_c(\T)$, 
the restriction to total \HT-inter\-pretations in $C_s(\T)$, 
respectively in $E_s(\T)$; 
\item $C_a(\T)$ 
the set of there-closed \HT-interpretations of the form 
$(\emptyset,Y)$ in $C_s(\T)$, and 
by  
$E_a(\T)$ 
the set of total-closed \HT-interpretations in $E_s(\T)$
(i.e., equilibrium models); 
\item $E_u(\T)$ 
the set of closed \HT-interpretations in 
$E_s(\T)$.
\end{itemize}
\end{definition}

By means of the above sets of \HT-countermodels, respectively equivalence 
interpretations, equivalences of propositional theories can be characterized 
as follows.

\begin{corollary}
Given two propositional theories $\T_1$ and $\T_2$, 
the following propositions are equivalent for $e\in\{c,a,s,u\}$:
\begin{eqnarray*}
\textrm{(1)}\ \ \T_1\equive\T_2; \ \ \ \ & 
\textrm{(2)}\ \  C_e(\T_1)=C_e(\T_2); \ \ \ \ &
\textrm{(3)}\ \  E_e(\T_1)=E_e(\T_2).
\end{eqnarray*}
\end{corollary}

\begin{example}\label{ex:infctd2}{\rm
In our running example,  $C_u(\T_1)\neq C_u(\T_2)$, as well as 
$E_u(\T_1)\neq E_u(\T_2)$, by the remarks
on non-total \HT-interpretations in Example~\ref{ex:infctd1}.
\qed}\end{example}


Since equivalence interpretations do not encompass total \HT-countermodels, we 
attempt a direct characterization in \HT. 

\begin{lemma}\label{lemm:total} 
For any \HT-interpretation $(X,Y)$ of signature $\sig$ and 
$\tau_{\epsilon}=\{\neg\neg a \IMP a\mid a \in \sig\}$, it holds that 
$(X,Y)\models  \tau_{\epsilon}$ iff $X=Y$.
\end{lemma}

\begin{proof}
$(X,Y)\models \tau_{\epsilon}$ for all $a \in \sig$ iff 
$(X,Y)\models \neg\neg a \IMP a$ for all $a \in \sig$ iff, for every 
$a \in \sig$, it holds that  $(X,Y)\not\models \neg\neg a$ or 
$(X,Y)\models a$, and $Y\models \neg\neg a \IMP a$. 
The latter is a tautology, and 
$(X,Y)\not\models \neg\neg a$ iff $a\not\in Y$. We conclude that 
$(X,Y)\models \tau_{\epsilon}$ iff $(X,Y)\models a$ for all $a \in Y$, i.e., 
iff $X=Y$.
\end{proof}

By means of this lemma, we can use formulas of the form $\neg\neg a \IMP a$ 
to ensure for a given formula $\F$ of $\T$ that if $(X,Y)\models\F$ then  
$X=Y$, i.e., that the \HT-interpretation is total.

\begin{proposition}\label{prop:char}
Let $M$ be an \HT-interpretation over $\sig$. Then, 
$M\in E_s(\T)$ for a theory $\T$ 
iff $M\models \T_\F$ for some $\F\in\T$, where 
$\T_\F=\{\neg\neg \G\mid \G\in\T\} \cup \{\F\IMP (\neg\neg a \IMP a)\mid a\in\sig\}$.
\end{proposition}

\begin{proof}
For the only-if direction, assume $(X,Y)$ is an equivalence interpretation of $\T$. Then 
$Y\models \G$ for all $\G\in\T$ and therefore  $(X,Y)\models \neg\neg \G$ for all 
$\G\in\T$. If $X=Y$, then by Lemma~\ref{lemm:total}, $(X,Y)$ also satisfies 
$\neg\neg a \IMP a$ 
for all $a\in\sig$. In this case, $(X,Y)\models \T_\F$ for all $\F\in\T$. 
We continue with the case where $X\subset Y$. Then, $(X,Y)$ is a 
here-countermodel of $\T$, i.e., there exists $\F\in\T$ such that 
$(X,Y)\not\models \F$.
This implies that $(X,Y)\models \F\IMP(\neg\neg a \IMP a)$ for all $a\in\sig$, i.e.,  
$(X,Y)\models \T_\F$. This proves 
the claim for $X\subset Y$.

For the if direction, assume that 
$(X,Y)\models \T_\F$ for some $\F\in\T$. Then, 
$(X,Y)\models \neg\neg \G$ for all $\G\in\T$, which implies $Y\models \G$ for all 
$\G\in\T$. Consequently, $(X,Y)$ is an equivalence interpretation of $\T$ if $X=Y$. 
If $X\subset Y$, we conclude that $(X,Y)$ does not satisfy $\neg\neg a \IMP a$ for some 
$a\in\sig$ by Lemma~\ref{lemm:total}. However, $(X,Y)\models \T_\F$ for some 
$\F\in\T$, hence $(X,Y)\models \F\IMP (\neg\neg a \IMP a)$ for all $a\in\sig$. Therefore, 
$(X,Y)\not\models \F$ must hold for some $\F\in\T$. This proves, since $X\subset Y$,  
that $(X,Y)$ is a here-countermodel of $\T$, i.e., an equivalence interpretation of $\T$.
\end{proof}

For infinite propositional theories, we thus end up with a characterization of 
equivalence interpretations as the union of the \HT-models of an infinite 
number of 
(infinite) theories. At least for finite theories, however, a characterization 
in terms of a (finite) theory is obtained (even for a potentially extended 
infinite signature).

If $\sig'\supset \sig$ and $M=(X,Y)$ is an \HT-interpretation over $\sig'$, 
then $M|_\sig$ denotes the restriction of 
$M$ to $\sig$:  $M|_\sig=(X|_\sig,Y|_\sig)$. The restriction is 
\emph{totality preserving}, if $X\subset Y$ implies $X|_\sig\subset Y|_\sig$.

\begin{proposition}\label{prop:submod}
Let $\T$ be a theory over $\sig$, let $\sig'\supset \sig$, and let  
$M$ an \HT-interpretation over $\sig'$ such that $M|_\sig$ is totality 
preserving. Then,  
$M\in C_s(\T)$ implies $M|_\sig\in C_s(\T)$.
\end{proposition}

\begin{proof}
Let $M=(X',Y')$, $M|_\sig=(X,Y)$, and assume $M\not\models\T$. 
First, suppose  $M$ is total, hence, $Y'\not\models \T$. 
Then, $Y\not\models \T$, because otherwise 
$Y'\models \T$ would hold, since $\T$ is over $\sig$.
This proves the claim for total \HT-countermodels, and since \HT\ 
countermodels are total-closed, for any \HT-countermodel $M=(X',Y')$, 
such that $Y'\not\models \T$. 

We continue with the case that $Y'\models \T$. Then $X'\subset Y'$ 
holds, which means that $M$ is an equivalence interpretation of $\T$. Therefore, 
$M\not\models\F$ for some $\F\in\T$. Additionally, $M\models \neg\neg \G$ 
for all $\G\in\T$ (recall that $Y'\models \T$). 
This implies $M\models\T_\F$, where  
$\T_\F=\{\neg\neg \G\mid \G\in\T\} \cup \{\F\IMP (\neg\neg a\IMP a)\mid a\in\sig\}$.
Therefore, $M|_\sig\models\T_\F$, i.e., $M|_\sig$ is an equivalence interpretation 
of $\T$. Since the restriction is totality preserving, $M|_\sig$ is non-total. 
This proves $M|_\sig\not\models\T$.
\end{proof}

This eventually enables the characterization of the \HT-countermodels of a 
finite theory by another finite theory, as stated in the next result.

\begin{theorem}\label{theo:tchar}
Let $\T$ be a finite theory over $\sig$, and let $M$ be an \HT-interpretation. 
Then, $M\in E_s(\T)$ 
iff $M|_\sig\models \bigvee_{\F\in\T}\bigwedge_{\G\in\T_\F} \G$, 
and $M|_\sig$ is totality preserving.
\end{theorem}

\begin{proof}
For the only-if direction let $M\in E_s(\T)$. If $M$ is total then 
$M|_\sig$ is total and $M\models\T$ implies $M|_\sig\models\T$. Hence,  
$M|_\sig\in E_s(\T)$ and 
$M|_\sig\models \bigvee_{\F\in\T}\bigwedge_{\G\in\T_\F} \G$.
So let $M$ be non-total. 
We show that $M|_\sig$ is totality-preserving. Towards a contradiction 
assume the contrary. Then, $M|_\sig$ is total. From $Y\models\T$ we 
conclude $Y|_\sig\models\T$ and the same for $X|_\sig$ by $X|_\sig=Y|_\sig$. 
Because $\T$ is over $\sig$, $X\models\T$ follows, hence $M\models\T$, 
which is a contradiction.
Thus, $M|_\sig$ is totality-preserving. Then $M|_\sig$ is also non-total 
and in $C_s(\T)$. Therefore $M|_\sig\in E_s(\T)$, which
implies $M|_\sig\models \bigvee_{\F\in\T}\bigwedge_{\G\in\T_\F} \G$.

For the if direction, consider any \HT-interpretation $M$ 
such that 
$M|_\sig$ satisfies the theory $\bigvee_{\F\in\T}\bigwedge_{\G\in\T_\F} \G$ 
and $M|_\sig$ is 
totality preserving. If $M$ is total then $M|_\sig$ is total and 
$M|_\sig\models\T$, which implies $M\models\T$, since $\T$ is over $\sig$.
If $M$ is non-total then $M|_\sig$ is non-total and $M|_\sig\not\models\T$,
which implies $M\not\models\T$.
\end{proof}

\begin{example}\label{ex:finrep}{\rm
Let $\T =\{a\}$ over $\sig=\{a\}$ and recall what the proof of 
Proposition~\ref{prop:nondual} established: There is no theory $\T'$ such 
that $(X,Y)$ is an \HT-model of $\T'$ iff it is an \HT-countermodel of $\T$. 
According to Theorem~\ref{theo:tchar} however, we can characterize $E_s(\T)$
by means of totality-preserving \HT-models 
of the theory $\T'=\{\neg\neg\, a \wedge (a\IMP(\neg\neg a\IMP a))\}$.
Consider any \HT-interpretation $(X,Y)$ over $\sig'\supset\sig$. 
It is easily verified that $(X,Y)\models \T'$ iff $a\in Y$. If additionally 
$a\in X$ and $X\subset Y$, then $(X|_\sig,Y|_\sig)$ is not totality preserving.
Thus, $(X,Y)$ is a totality-preserving \HT-model of $\T'$ iff $a\in Y$ and 
either $X=Y$ or $a\not\in X$. These interpretations respectively correspond to 
the total models and the here-countermodels, i.e., the equivalence interpretations 
of $\T$ over $\sig'$. 
\qed}
\end{example}

\subsection{Relativized Hyperequivalence for Propositional Theories}

We now turn to the notion of relativized hyperequivalence. The term 
`hyperequivalence' has been coined in the context of ASP, as a general 
expression for different forms of equivalence, which guarantee that the 
semantics is preserved under the addition of arbitrary programs (called 
\emph{contexts}) from a particular class of 
programs~\cite{trus-wolt-08b}. Relativized 
hyperequivalence emanates from the study of relativized notions of 
equivalence by restricting contexts to particular alphabets (see 
e.g.,~\cite{eite-etal-07a,pear-etal-07}). It has been 
generalized to the setting, where possibly different alphabets are used to 
restrict the head atoms and the body atoms allowed to appear in context 
rules~\cite{wolt-08}.

While up to now relativized hyperequivalence has only been studied for finite 
programs, we aim at a generalization of relativized hyperequivalence for 
propositional theories under the answer-set semantics, without any finiteness 
restrictions. For this purpose, we first generalize the notions of `head atom' 
and `body atom' for theories.

The occurrence of an atom $a$ in a formula $\F$ is called 
\emph{positive} if $\F$ is implication free, if $a$ occurs in the 
consequent of an implication in $\F$, or if $\F$ is of the form 
$(\F_1\IMP\F_2)\IMP\F_3$ and $a$ occors in $\F_1$. 
An occurrence of $a$ is called 
\emph{negative} if $a$ occurs in the antecedent of an implication. 
The notion of positive and negative occurrence is extended to (sub-)formulas 
in the obvious way.
Note that any occurrence under negation therefore is a negative occurrence, 
and that the occurrence of an atom or subformula may be both positive and 
negative, for instance the occurrence of $b$ in $a\IMP (b\IMP \bot)$, 
viz.~$a\IMP \neg b$.

A propositional theory $\T$ over $\Ap\cup\An$, where $\Ap$ and $\An$ are sets 
of propositional variables, 
is called an \emph{$\Ap$-$\An$-theory} if every formula in $\T$ has positive 
occurrences of atoms from $\Ap$, and negative occurrences of atoms from $\An$, 
only. Note that $\bot$ is always allowed to appear both, positively and 
negatively. An $\Ap$-$\An$-theory is called \emph{extended}, if additionally 
factual formulas over $\Ap$ 
are permitted. 

By means of these notions, relativized hyperequivalence for propositional 
theories can be expressed as follows, which is a proper generalization of 
the logic programming setting.

\begin{definition}\label{def:hyperequiv}
Two propositional theories  $\T_1, \T_2$ over $\sig$ are called
\emph{relativized hyperequivalent wrt.~$\Ap$ and $\An$}, symbolically 
$\T_1\equivh\T_2$, iff for any $\Ap$-$\An$-theory $\T$ over 
$\sig'\supseteq\sig$, $\T_1\cup \T$ and  $\T_2\cup \T$ are answer-set 
equivalent.
\end{definition}

Towards a characterization of relativized hyperequivalence, our goal is to 
follow the same methodology that we used  to characterize uniform 
equivalence, i.e., resorting to \HT-countermodels and respective closure 
conditions. However, while in the logic programming setting such closure 
conditions may be obtained from certain monotonicity properties of the 
program reduct, we first have to establish corresponding properties for 
theories. A first property in this respect is the following. Note that 
although the next result is stated for extended $\Ap$-$\An$-theories (for 
reasons which will become clear later), it trivially also holds for any 
(non-extended) $\Ap$-$\An$-theory.

\begin{proposition}\label{prop:hyp-eval}
Consider an extended propositional $\Ap$-$\An$-theory $\T$, 
and an \HT-interpretation $(X,Y)$. Then, 
$(X,Y)\models \T$ implies 
$(X',Y)\models \T$, 
for all $X'\subseteq Y$ such that 
$X|_{\Ap} \subseteq X'|_{\Ap}$ and $X'|_{\An} \subseteq X|_{\An}$.
\end{proposition}

\begin{proof}
Consider any $\Ap$-$\An$-formula $\F$ in $\T$, i.e., any formula that has 
positive occurrences of atoms from $\Ap$, and negative occurrences of atoms 
from $\An$, only.
We show by induction on the formula structure  of $\F$, that for all 
$X'\subseteq Y$ such that 
$X|_{\Ap} \subseteq X'|_{\Ap}$ and  $X'|_{\An} \subseteq X|_{\An}$: 
\begin{description}
\item[(a)] $(X,Y)\models\F$ implies $(X',Y)\models\F$ if $\F$ is a positive 
occurrence; and 
\item[(b)] $(X,Y)\not\models\F$ implies $(X',Y)\models\F$ if $\F$ is a 
negative occurrence.
\end{description}

For the base case, consider any atomic formula $\F$, and suppose first that 
(a) the occurrence of $\F$ is a positive occurrence. Then, $(X,Y)\models\F$ 
implies that $\F$ is not $\bot$, and thus is an atom $a$ from $\Ap$ such that 
$a\in X$. Since $X|_{\Ap} \subseteq X'|_{\Ap}$ for all $X'$ under 
consideration, we conclude that $a\in X'$. Hence,  $(X',Y)\models\F$.
Suppose (b) $\F$ is a negative occurrence. 
If $(X,Y)\not\models\F$, then either $\F$ is $\bot$, and 
$(X',Y)\not\models\F$ follows trivially. Otherwise, $\F$ is an atom $b$ from 
$\An$, such that $b\not\in X|_{\An}$. Since 
$X'|_{\An} \subseteq X|_{\An}$ 
for all $X'$ under consideration, we conclude that $b\not\in X'$, i.e., 
$(X',Y)\not\models\F$. This proves (a) and (b) for atomic formulas.

For the induction step, assume that (a) and (b) hold for any 
$\Ap$-$\An$-formula of connective nesting depth $n-1$, and let $\F$ be a 
formula of connective nesting depth $n$.
Consider the case where $\F$ is of the form $\F_1\AND\F_2$, respectively 
$\F_1\OR\F_2$.
If $\F$ is a positive occurrence (a), then so are $\F_1$ and $\F_2$, both of 
connective nesting depth depth $n-1$. 
From $(X,Y)\models\F$ we conclude $(X,Y)\models\F_1$ and (or) 
$(X,Y)\models\F_2$. The induction hypothesis applies, proving 
$(X',Y)\models\F_1$ and (or) 
$(X',Y)\models\F_2$, for all $X'\subseteq Y$ such that 
$X|_{\Ap} \subseteq X'|_{\Ap}$ and $X'|_{\An} = X|_{\An}$,
i.e., $(X',Y)\models\F$ for all $X'$ under consideration.
In case $\F$ is a negative occurrence (b), then so are $\F_1$ and $\F_2$, 
both of connective nesting depth $n-1$. Then, 
$(X,Y)\not\models\F$ implies
$(X,Y)\not\models\F_1$ or (and) $(X,Y)\not\models\F_2$, and the same holds for 
any $(X',Y)$ under consideration by induction hypothesis. This proves 
$(X,Y)\not\models\F$ implies $(X',Y)\models\F$.

Finally, let  $\F$ be of the form $\F_1\IMP\F_2$. Then, independent of 
whether $\F$ occurs positively or negatively,  $\F_1$ is a negative occurrence 
and $\F_2$ is a positive occurrence, both of connective nesting depth $n-1$. 
First, suppose that $\F$ is a positive occurrence (a), as well as that 
$(X,Y)\models\F$. 
Towards a contradiction assume that there exists $X'\subseteq Y$ such that 
$X|_{\Ap} \subseteq X'|_{\Ap}$, 
$X'|_{\An} \subseteq X|_{\An}$, and 
$(X',Y)\not\models \F$. Since $(X,Y)\models\F$ implies that 
$Y\models\F$, we conclude that both, $(X',Y)\models \F_1$ and 
$(X',Y)\not\models\F_2$, hold. From the latter, 
since $\F_2$ is a positive occurrence of connective nesting depth $n-1$, 
it follows that $(X,Y)\not\models\F_2$ (otherwise by induction hypothesis (a) 
$(X',Y)\models\F_2$). 
This implies $(X,Y)\not\models\F_1$ since $(X,Y)\models\F$. However, 
$\F_1$ is a negative occurrence of connective nesting depth $n-1$, thus by 
induction hypothesis (b) we conclude that $(X',Y)\not\models \F_1$, a 
contradiction. Therefore, $(X',Y)\models\F$ for all  $X'$ under consideration, 
which proves (a). 
For 
(b), let $\F$ be a negative occurrence and suppose 
$(X,Y)\not\models\F$.
If $Y\not\models\F$, then also $(X',Y)\not\models\F$ for all $X'$ under 
consideration. In case $Y\models\F$, we conclude that $(X,Y)\models \F_1$ and 
$(X,Y)\not\models\F_2$. Since  $\F$ is a negative occurrence, not only $\F_1$ 
but also $\F_2$ is a negative occurrence, both of connective nesting depth 
$n-1$. Therefore, by induction hypothesis (b) we conclude that 
$(X',Y)\not\models \F_2$. Moreover, also because $\F$ is a negative occurrence,
$\F_1$ is a positive occurrence as well. Hence, by induction hypothesis (a) we 
conclude $(X',Y)\models \F_1$ from $(X,Y)\models \F_1$, 
viz.~$(X',Y)\not\models \F$, for all $X'$ under 
consideration. This concludes the inductive argument and proves (a) and (b) for 
$\Ap$-$\An$-formulas of arbitrary connective nesting. 

Next, we turn to factual formulas $\G$ in $\T$, 
and prove by induction on the formula structure of $\G$, that
\begin{description}
\item[(c)] $(X,Y)\models\G$ implies $(X',Y)\models\G$, for all $X'\subseteq Y$ 
such that $X|_{\Ap} \subseteq X'|_{\Ap}$ and  $X'|_{\An} \subseteq X|_{\An}$; and 
\item[(d)] $(Y,Y)\not\models\G$ implies $(X',Y)\not\models\G$, for all 
$X'\subseteq Y$.
\end{description}

For the base case, consider any atomic formula $\G$, and suppose first that 
(c) $(X,Y)\models\G$. Then, $\G$ is not $\bot$, but an atom $a$ from 
$\Ap$
such that $a\in X$. Since $X|_{\Ap} \subseteq X'|_{\Ap}$ for all 
$X'$ such that $X|_{\Ap} \subseteq X'|_{\Ap}$ and $X'|_{\An} \subseteq X|_{\An}$, 
we conclude that $a\in X'$. Hence,  $(X',Y)\models\G$.
For (d), assume $(Y,Y)\not\models\G$. 
Then $\G$ is $\bot$ or $\G$ is a an atom not in $Y$. 
In the former case, $(X',Y)\not\models\G$ follows trivially for all 
$X'\subseteq Y$. In the latter case, the atom also cannot be a member of any 
$X'$ such that $X'\subseteq Y$. Therefore, $(X',Y)\not\models\G$, for all 
$X'\subseteq Y$. This proves (c) and (d) for atomic formulas.

For the induction step, assume that (c) and (d) hold for any factual 
formula of connective nesting depth $n-1$, and let $\G$ be a factual formula of 
connective nesting depth $n$.
Consider the case where $\G$ is of the form $\G_1\AND\G_2$, respectively 
$\G_1\OR\G_2$. Since $\G$ is factual, so are $\G_1$ and $\G_2$, both of 
connective nesting depth depth $n-1$.
In case (c), from $(X,Y)\models\G$ we conclude $(X,Y)\models\G_1$ and (or) 
$(X,Y)\models\G_2$. The induction hypothesis applies, proving 
$(X',Y)\models\G_1$ and (or) 
$(X',Y)\models\G_2$, for all $X'\subseteq Y$ such that 
$X|_{\Ap} \subseteq X'|_{\Ap}$ and 
$X'|_{\An} \subseteq X|_{\An}$, 
i.e., $(X',Y)\models\G$ for all $X'$ under consideration.
Assume (d), i.e., $(Y,Y)\not\models\G$.  As a consequence, 
$(Y,Y)\not\models\G_1$ or (and) $(Y,Y)\not\models\G_2$, hence by induction 
hypothesis, for all $X'\subseteq Y$, it holds that $(X',Y)\not\models\G_1$ 
or (and) $(X',Y)\not\models\G_2$. 
Therefore, $(X',Y)\not\models\G$, for all $X'\subseteq Y$.

Finally, let  $\G$ be of the form $\G_1\IMP\bot$. Then, $\G_1$ is factual and 
of connective nesting depth depth $n-1$.
In case (c), if $(X,Y)\models\G$, then $Y\models\G$, hence 
$Y\not\models\G_1$, i.e., $(Y,Y)\not\models\G_1$ and by induction 
hypothesis (d), the same holds for any $(X',Y)$ such that $X'\subseteq Y$. 
Thus, in particular for 
$X'\subseteq Y$ such that $X|_{\Ap} \subseteq X'|_{\Ap}$ and 
$X'|_{\An} \subseteq X|_{\An}$, it follows that $(X',Y)\not\models\G_1$.
Moreover, $Y\models\G$, and therefore $(X',Y)\models\G\IMP\bot$, 
for all $X'\subseteq Y$ such that $X|_{\Ap} \subseteq X'|_{\Ap}$ and 
$X'|_{\An} \subseteq X|_{\An}$.
For (d), assume $(Y,Y)\not\models\G$. Consequently $Y\not\models\G$, and this 
implies $(X',Y)\not\models\G$, for all $X'\subseteq Y$. This concludes the 
inductive argument and proves (c) and (d) for factual formulas over 
$\Ap$ of arbitrary connective nesting. 

Concerning the claim of the proposition, since $(X,Y)\models \T$ implies 
$(X,Y)\models \F$ and $(X,Y)\models \G$, for every $\Ap$-$\An$-formula $\F$ in 
$\T$ and every factual formula $\G$ in $\T$, we conclude that $(X',Y)\models \F$ 
and $(X',Y)\models \G$, for all $X'\subseteq Y$ such that 
$X|_{\Ap} \subseteq X'|_{\Ap}$ and $X'|_{\An} \subseteq X|_{\An}$. 
This proves $(X',Y)\models \T$, for all $X'$ under consideration.
\end{proof}

Complementary to this result, given a total \HT-model of an (extended) 
$\Ap$-$\An$-theory, we can infer its satisfaction for the following class of 
non-total \HT-interpretations.  

\begin{proposition}\label{prop:hyp-tot}
Consider an extended propositional $\Ap$-$\An$-theory $\T$, 
and a total \HT-interpretation $(Y,Y)$. Then, 
$(Y,Y)\models \T$ implies $(X',Y)\models \T$, 
for all $X'\subseteq Y$ such that 
$X'|_{\Ap} = Y|_{\Ap}$. 
\end{proposition}

\begin{proof}
Consider any $\Ap$-$\An$-formula $\F$ in $\T$, i.e., any formula that has 
positive occurrences of atoms from $\Ap$, and negative occurrences of atoms 
from $\An$, only.
We show by induction on the formula structure  of $\F$, that for all 
$X'\subseteq Y$ such that 
$X'|_{\Ap} = Y|_{\Ap}$: 
\begin{description}
\item[(a)] $(Y,Y)\models\F$ implies $(X',Y)\models\F$ if $\F$ is a positive 
occurrence; and
\item[(b)] $(Y,Y)\not\models\F$ implies $(X',Y)\not\models\F$ if $\F$ is a 
negative occurrence. 
\end{description}

For the base case, consider any atomic formula $\F$, and suppose first (a)
that $\F$ is a positive occurrence such that $(Y,Y)\models\F$. 
Then $\F$ is not $\bot$, and thus is an atom $a$ from 
$\Ap$ such that $a\in Y$. Since $X'|_{\Ap} = Y|_{\Ap}$ for all $X'$ under 
consideration, we conclude that $a\in X'$. Hence,  $(X',Y)\models\F$.
Suppose (b) $\F$ is a negative occurrence. If $(Y,Y)\not\models\F$, then 
$\F$ is either $\bot$, or an atom $b$ from $\An$, such that 
$b\not\in Y$. Since $X'\subseteq Y$ implies $X'|_{\An} \subseteq Y|_{\An}$ 
for all $X'$ under consideration, we conclude that $b\not\in X'$. 
Hence,  $(X',Y)\not\models\F$.

For the induction step, assume that (a) and (b) hold for any 
$\Ap$-$\An$-formula of connective nesting depth $n-1$, and let $\F$ be a 
formula of connective nesting depth $n$.
Consider the case where $\F$ is of the form $\F_1\AND\F_2$, respectively 
$\F_1\OR\F_2$.
If $\F$ is a positive occurrence (a), then so are $\F_1$ and $\F_2$, both of 
connective nesting depth depth $n-1$. 
From $(Y,Y)\models\F$ we conclude $(Y,Y)\models\F_1$ and (or) 
$(Y,Y)\models\F_2$. The induction hypothesis applies, proving 
$(X',Y)\models\F_1$ and (or) 
$(X',Y)\models\F_2$, for all $X'\subseteq Y$ such that 
$X'|_{\Ap} = Y|_{\Ap}$, 
i.e., $(X',Y)\models\F$ for all $X'$ under consideration.
In case $\F$ is a negative occurrence (b), then so are $\F_1$ and $\F_2$, 
both of connective nesting depth $n-1$. Then, $(Y,Y)\not\models\F$ implies 
$(Y,Y)\not\models\F_1$ or (and) $(Y,Y)\not\models\F_2$, and the same holds 
for any $(X',Y)$ under consideration by induction hypothesis. This proves 
$(X',Y)\not\models\F$.
Finally, let  $\F$ be of the form $\F_1\IMP\F_2$. Then, independent of 
whether $\F$ occurs positively or negatively,  $\F_1$ is a negative occurrence 
and $\F_2$ is a positive occurrence, both of connective nesting depth $n-1$. 
First, suppose 
$(Y,Y)\models\F$. 
Towards a contradiction assume that there exists $X'\subseteq Y$ such that 
$X'|_{\Ap} = Y|_{\Ap}$ and $(X',Y)\not\models\F$.
Since $(Y,Y)\models\F$ implies that 
$Y\models\F$, we conclude that both, $(X',Y)\models \F_1$ and 
$(X',Y)\not\models\F_2$, hold. From the latter, 
since $\F_2$ is a positive occurrence of connective nesting depth $n-1$, 
it follows that $(Y,Y)\not\models\F_2$ (otherwise by induction hypothesis (a) 
$(X',Y)\models\F_2$). 
This implies $(Y,Y)\not\models\F_1$ since $(Y,Y)\models\F$. However, 
$\F_1$ is a negative occurrence of connective nesting depth $n-1$, thus by 
induction hypothesis (b) we conclude that $(X',Y)\not\models \F_1$, a 
contradiction. Therefore, $(X',Y)\models\F$ for all  $X'$ under consideration, 
which proves (a).
For (b), let $\F$ be a negative occurrence and suppose $(Y,Y)\not\models\F$.
Then $Y\not\models\F$, hence also $(X',Y)\not\models\F$ for all $X'$ under 
consideration. This concludes the inductive argument and proves (a) and (b) for 
$\Ap$-$\An$-formulas of arbitrary connective nesting. 

Next, we turn to factual formulas $\G$ in $\T$, 
and prove by induction on the formula structure of $\G$, that $(Y,Y)\models\G$ 
implies $(X',Y)\models\G$, for all $X'\subseteq Y$ such that 
$X'|_{\Ap} = Y|_{\Ap}$.

For the base case, consider any atomic formula $\G$, and suppose that 
$(Y,Y)\models\G$. Then, $\G$ is not $\bot$, but an atom $a$ from $\Ap$ 
such that $a\in Y$. Since $X'|_{\Ap} = Y|_{\Ap}$ for all $X'$ under 
consideration, we conclude that $a\in X'$. Hence,  $(X',Y)\models\G$, 
for all $X'\subseteq Y$ such that $X'|_{\Ap} = Y|_{\Ap}$.

For the induction step, assume that the claim holds for any factual 
formula of connective nesting depth $n-1$, and let $\G$ be a factual formula of 
connective nesting depth $n$.
Consider the case where $\G$ is of the form $\G_1\AND\G_2$, respectively 
$\G_1\OR\G_2$. Since $\G$ is factual, so are $\G_1$ and $\G_2$, both of 
connective nesting depth depth $n-1$.
From $(Y,Y)\models\G$ we conclude $(Y,Y)\models\G_1$ and (or) 
$(Y,Y)\models\G_2$. The induction hypothesis applies, proving 
$(X',Y)\models\G_1$ and (or) 
$(X',Y)\models\G_2$, for all $X'\subseteq Y$ such that $X'|_{\Ap} = Y|_{\Ap}$,
i.e., $(X',Y)\models\G$ for all $X'$ under consideration.
Finally, let  $\G$ be of the form $\G_1\IMP\bot$. Then, $\G_1$ is factual and 
of connective nesting depth depth $n-1$.
If $(Y,Y)\models\G$, then $Y\models\G$, hence 
$Y\not\models\G_1$, i.e., $(Y,Y)\not\models\G_1$ and by Case~(d) in the proof of 
Proposition~\ref{prop:hyp-eval}, the same holds for any $(X',Y)$ such that 
$X'\subseteq Y$. Thus, in particular for 
$X'\subseteq Y$ such that $X'|_{\Ap} = Y|_{\Ap}$, it follows that 
$(X',Y)\not\models\G_1$.
Moreover, $Y\models\G$, and therefore $(X',Y)\models\G\IMP\bot$, 
for all $X'\subseteq Y$ such that $X'|_{\Ap} = Y|_{\Ap}$.
This concludes the inductive argument and proves the claim for factual formulas 
over 
$\Ap$ of arbitrary connective nesting. 

Concerning the claim of the proposition, since $(Y,Y)\models \T$ implies 
$(Y,Y)\models \F$ and $(Y,Y)\models \G$, for every $\Ap$-$\An$-formula $\F$ in 
$\T$ and every factual formula $\G$ in $\T$, we conclude that $(X',Y)\models \F$ 
and $(X',Y)\models \G$, for all $X'\subseteq Y$ such that 
$X'|_{\Ap} = Y|_{\Ap}$. 
This proves $(X',Y)\models \T$, for all $X'$ under consideration.
\end{proof}

Having established these properties of $\Ap$-$\An$-theories, we can state 
respective closure conditions for \HT-interpretations referring to 
countermodels, or which we consider more convenient here, referring to 
equivalence interpretations.

\begin{definition}\label{def:hyperclosure}
Given a propositional theory $\T$ over $\sig$, sets of propositional variables 
$\Ap\subseteq\sig'$, $\An\subseteq\sig'$, $\sig'\supseteq\sig$, and an 
\HT-interpretation $(X,Y)$, we say that
\begin{itemize}
\item $(Y,Y)$ is \emph{$\Ap$-total} iff $(Y|_{\Ap},Y)$ is closed in $E_s(\T)$;
\item 
  $(X,Y)$ is \emph{$\Ap$-closed} in $E_s(\T)$ iff 
   $(X',Y)\in E_s(\T)$, for all $X'\subseteq Y$ such that
       $X|_{\Ap}\subseteq X'|_{\Ap}$ and $X'|_{\An} \subseteq X|_{\An}$. 
\end{itemize}
\end{definition}

With these concepts, a semantic characterization of relativized 
hyperequivalence for propositional theories can be established by means of 
the following characteristic equivalence interpretations.

\begin{definition}\label{def:hypercmodel}
An \HT-interpretation $(X,Y)$ is an \emph{\HT-hyperequivalence 
interpretation wrt.~$\Ap$ and $\An$} of a propositional theory $\T$
iff $(Y,Y)$ is $\Ap$-total and 
there exists an \HT-interpretation $(X',Y)$ such that 
$X=X'|_{\Ap\cup\An}$ and $(X',Y)$ is $\Ap$-closed in $E_s(\T)$.

\noindent
The set of \HT-hyperequivalence interpretations wrt.~$\Ap$ and $\An$ 
of a propositional theory $\T$ is denoted by $\Eh{\T}$.
\end{definition}

This definition intuitively generalizes the characterization of 
Woltran~\citeyear{wolt-08} for the logic programming setting to propositional 
theories. Note however, that rather than resorting to \HT-models and a 
maximality criterion, the above definition refers to equivalence 
interpretations (i.e., \HT-countermodels in case of non-totality) and 
respective closure conditions. As in the case of uniform equivalence, this 
not only simplifies the definition, but also avoids difficulties in infinite 
settings. The next result establishes that \HT-hyperequivalence 
interpretations precisely characterize relativized hyperequivalence.

\begin{theorem}\label{theo:hyper}
Two propositional theories  $\T_1, \T_2$ 
are relativized hyperequivalent wrt.~$\Ap$ and $\An$ if and only if 
they coincide on their \HT-hyperequivalence interpretations 
wrt.~$\Ap$ and $\An$, 
symbolically  $\T_1\equivh\T_2$ iff $\Eh{\T_1}=\Eh{\T_2}$.
\end{theorem}

\begin{proof}
In the following, we will use the following notational simplification:
For any set of atoms $X$, we write 
$X_+$ for $X|_{\Ap}$, and $X_-$ for $X|_{\An}$.

For the only-if direction suppose $\T_1\equivh\T_2$ and towards a 
contradiction assume that $\Eh{\T_1}\neq\Eh{\T_2}$. W.l.o.g.~let 
$(X,Y)\in\Eh{\T_1}$ and $(X,Y)\not\in\Eh{\T_2}$ (the other case is symmetric).
Note that $(X,Y)\in\Eh{\T_1}$ implies that $(Y,Y)$ is $\Ap$-total, i.e., 
$(Y_+, Y)$ is closed in $E_s(\T_1)$. 
This implies that $(Y_+, Y)$ is in  $\Eh{\T_1}$. 
Suppose 
$(Y_+, Y)$ is not in $\Eh{\T_2}$. Then, either 
$(Y,Y)\not\models \T_2$, or there exists $Y_+\subseteq X'\subset Y$
such that $(X',Y)\models \T_2$. Let $\T=Y_+$ and observe that in both 
cases $Y$ is not an answer set of $\T_2\cup\T$. In the former case because 
$(Y,Y)\not\models \T_2\cup\T$, in the latter because $X'\subset Y$ and
$(X',Y)\models \T_2\cup\T$ (note that $(X',Y)\models \T$ by 
Proposition~\ref{prop:hyp-tot}).
However, $Y$ is an answer set of $\T_1\cup\T$.
Indeed, $(Y_+, Y)$ is closed in $E_s(\T_1)$. 
And for any $X'\subset Y$ such that $Y_+\not\subseteq X'_+$, 
obviously $(X',Y)$ is a non-total \HT-countermodel of $\T$. 
Consequently $(Y,Y)$ is total-closed in $E_s(\T_1\cup\T)$. 
Because $\T$ is an $\Ap$-$\An$-theory, this contradicts $\T_1\equivh\T_2$.
Thus, we conclude that 
$(Y_+, Y)\in\Eh{\T_2}$.
Note that therefore $(Y,Y)$ is $\Ap$-total for $\T_2$, which implies that
$(Y|_A, Y)$ is in  $\Eh{\T_2}$, hence $X\subset Y|_A$ and 
$X_+\subset Y_+$. 
Consider the following theory 
$\T=X_+\cup\{\alpha\IMP\beta \mid 
\alpha\in Y_-\setminus X_-, \beta\in Y_+\setminus X_+\}$.
We show that $Y$ is an answer set of $\T_1 \cup \T$. Obviously, $Y\models\T$ 
because $X_+\subset Y_+$ and 
$\beta\in Y$ for every $\beta\in Y_+\setminus X_+$. 
Therefore, 
$(Y,Y)\models \T_1 \cup \T$. Towards a contradiction, assume that there 
exists $X'\subset Y$ such that $(X',Y)\models \T_1 \cup \T$. From 
$(X',Y)\models \T$, we conclude that either $X'_+ = Y_+$, or that 
$X_+\subseteq X'_+ \subset Y_+$ and $X'_- \subseteq X_-$.
In both cases, $(X',Y)\not\models \T_1$. In the former case because 
$(Y,Y)$ is $\Ap$-total, i.e.,  $(Y_+, Y)$ is closed in $E_s(\T_1)$. 
In the latter case, it is a consequence of the fact that $(X,Y)\in\Eh{\T_1}$, 
which implies $(X',Y)\not\models \T_1$ by $\Ap$-closure. This contradicts our 
assumption concerning the existence of $X'\subset Y$ such that 
$(X',Y)\models \T_1 \cup \T$, and proves that $Y$ is an answer set of 
$\T_1 \cup \T$. However, $Y$ is not an answer set of $\T_2 \cup \T$.
To wit, since $(X,Y)\not\in\Eh{\T_2}$, there exists $X'\subset Y$ such that 
$X_+\subseteq X'_+$, $X'_- \subseteq X_-$, and 
$(X',Y)\models \T_2$. Moreover, $(X',Y)$ is an \HT-model of $\T$. Observe that 
$X'_- \subseteq X_-$ implies that $(X',Y)$ is an \HT-model of every formula of the 
form 
$\alpha\IMP\beta$ in $\T$.
Hence, $(X',Y)\models \T_2 \cup \T$, and since $X'\subset Y$, it follows that
$Y$ is not an answer set of $\T_2 \cup \T$. Note that $\T$ is an 
$\Ap$-$\An$-theory, which contradicts $\T_1\equivh\T_2$. This proves 
$\Eh{\T_1}=\Eh{\T_2}$.

For the if direction, suppose $\Eh{\T_1}=\Eh{\T_2}$ and towards a 
contradiction assume that $\T_1\nequivh\T_2$. W.l.o.g.~let $Y$ be an 
answer set of $\T_1\cup \T$ for some  $\Ap$-$\An$-theory $\T$, such that 
$Y$ is not an answer set of $\T_2\cup \T$ (the other case is symmetric).
Then, $(Y,Y)$ is an equivalence interpretation of both, $\T_1$ and $\T$, and 
$(Y_+, Y)$ is closed in $E_s(\T_1\cup\T)$, which implies 
(taking Proposition~\ref{prop:hyp-tot} into account)
that $(Y,Y)$ is $\Ap$-total for $\T_1$ and $(Y|_A, Y)$ is in  $\Eh{\T_1}$.
Therefore, $(Y|_A, Y)$ is also in $\Eh{\T_2}$, 
with the consequence that $(Y, Y)$ is in $E_s(\T_2)$, and thus 
$(Y, Y)\in E_s(\T_2\cup\T)$. Since by assumption $Y$ is not an answer set 
of $\T_2\cup\T$, there exists $X\subset Y$ such that 
$(X,Y)\not\in E_s(\T_2\cup\T)$, i.e., $(X, Y)\models \T_2\cup\T$.
Since $(Y|_A, Y)\in\Eh{\T_2}$, 
it holds that $X|_A\subset Y|_A$. Moreover, $X_+\subset Y_+$ due to 
$\Ap$-totality of $(Y,Y)$.
Clearly, $(X|_A, Y)$ is not in $\Eh{\T_2}$ as witnessed by $(X,Y)\models\T_2$, 
and thus $(X|_A, Y)\not\in\Eh{\T_1}$ since $\Eh{\T_1}=\Eh{\T_2}$. 
From $(X|_A, Y)\not\in\Eh{\T_1}$, we conclude that there exists $X'\subseteq Y$, 
such that $X_+\subseteq X'_+$, $X'_- \subseteq X_-$, 
and $(X', Y)\not\in E_s(\T_1)$, i.e., $X'\subset Y$ and $(X', Y)\models \T_1$. 
By Proposition~\ref{prop:hyp-eval}, $(X, Y)\models \T$ implies  
$(X', Y)\models \T$. 
Consequently, $(X', Y)\models \T_1\cup\T$, 
and since $X'\subset Y$, this contradicts 
our assumption that $Y$ is an answer set of $\T_1 \cup \T$, and 
proves $\T_1\equivh\T_2$.
\end{proof}

Like in the logic programming setting, the framework obtained by the 
consideration of relativized hyperequivalence interpretations provides 
a general unified characterization of semantic characterizations of 
equivalence notions. In other words, the notions of equivalence considered 
in the previous subsection are obtained as special cases. For this purpose, 
one needs to refer to the universal alphabet (respectively signature), denoted   
by $\at$, explicitely. Then, by definition, setting $\Ap=\An=\emptyset$ amounts 
to answer-set equivalence, $\Ap=\An=\at$ yields strong equivalence, and 
$\Ap=\at$, $\An=\emptyset$ characterizes uniform equivalence. 
The latter is not by definition but follows from two simple observations: every 
set of facts over $\at$ is a $\at$-$\emptyset$-theory, and every 
$\at$-$\emptyset$-theory is a factual theory modulo formulas of the form 
$\bot\IMP\phi$, which are tautologies in \HT.

\begin{corollary}\label{coro:hspec}
Given two propositional theories $\T_1$ and $\T_2$ over $\sig\subseteq\at$, 
the following propositions are equivalent for $e\in\{a,s,u\}$,  
$\Ap(a)=\An(a)=\emptyset$, $\Ap(s)=\An(s)=\at$, $\Ap(u)=\at$, 
and $\An(u)=\emptyset$:
\begin{eqnarray*}
\textrm{(1)}\ \ \T_1\equive\T_2; \ \ \ \ & 
\textrm{(2)}\ \  \T_1\ ^{\Ap(e)}_{\An(e)}\!\!\equiv\, \T_2.
\end{eqnarray*}
\end{corollary}

In 
these particular cases, not only the notions of equivalence but also the 
characteristic semantic structures coincide, i.e., relativized hyperequivalence 
interpretations coincide with the respective characteristic sets of equivalence 
interpretations. 

\begin{proposition}\label{prop:hstruct}
Let $\T$ be 
a propositional theory 
over $\sig\subseteq\at$, and 
let 
$e\in\{a,s,u\}$,  
$\Ap(a)=\An(a)=\emptyset$, $\Ap(s)=\An(s)=\at$, $\Ap(u)=\at$, 
and $\An(u)=\emptyset$.
Then, 
$$E_e(\T)= E^{\Ap(e)}_{\An(e)}(\T).$$
\end{proposition}

\begin{proof}
First consider answer-set equivalence, i.e., $e=a$ and $\Ap=\An=\emptyset$. 
Then for any \HT-interpretation $(X,Y)$, it holds that 
$(X,Y)\in \Eh{\T}=E^{\Ap(e)}_{\An(e)}(\T)$ 
iff $(\emptyset,Y)$ is there-closed in $E_s(\T)$ and $X=\emptyset$. The former 
follows from the first condition in Definition~\ref{def:hypercmodel} since 
$Y|_{\Ap}=\emptyset$, and the latter from the second condition in 
Definition~\ref{def:hypercmodel}, i.e., from the existence of an $X'$ such that 
$X=X'|_{\Ap}$ (since $X'|_\emptyset=\emptyset$ for any $X'$).
Note that $X=\emptyset$ and $(\emptyset,Y)$ there-closed in $E_s(\T)$ are 
exactly the requirements for $(X,Y)\in E_a(\T)$. This proves $(X,Y)\in \Eh{\T}$ 
iff $(X,Y)\in E_a(\T)$.

Turning to strong equivalence, let $e=s$ and $\Ap=\An=\at$.
Then for any \HT-interpretation $(X,Y)$ over $\at$, it holds that
$(X,Y)\in \Eh{\T}=E^{\Ap(e)}_{\An(e)}(\T)$ 
iff $(Y,Y)$ in $E_s(\T)$ and $(X,Y)$ in $E_s(\T)$. The former 
follows from the first condition in Definition~\ref{def:hypercmodel} since 
$Y|_\at=Y$, and the latter from the second condition in 
Definition~\ref{def:hypercmodel}, i.e., from the existence of an $X'$ such that 
$X=X'|_\at$ (which implies $X'=X$ since $X'|_\at=X'$ for any $X'$) and such that 
$X''\in E_s(\T)$ for all $X''\subseteq Y$ where $X'|_\at=X''|_\at$ (i.e., for 
$X''=X'=X$).
Note that  $(X,Y)\in E_s(\T)$ implies $(Y,Y)\in E_s(\T)$. Consequently, 
it holds that $(X,Y)\in \Eh{\T}$ iff $(X,Y)\in E_s(\T)$.

Eventually consider uniform equivalence, i.e., $e=u$, $\Ap=\at$, and 
$\An=\emptyset$. In this case, $(X,Y)\in \Eh{\T}=E^{\Ap(e)}_{\An(e)}(\T)$, 
for any \HT-interpretation $(X,Y)$ over $\at$, iff 
$(Y,Y)$ in $E_s(\T)$ and  $(X',Y)$ in $E_s(\T)$ for all 
$X\subseteq X''\subseteq Y$. The former 
follows from the first condition in Definition~\ref{def:hypercmodel} since 
$Y|_\at=Y$, and the latter from the second condition in 
Definition~\ref{def:hypercmodel}, i.e., from the existence of an $X'$ such that 
$X=X'|_\at$ (which implies $X'=X$ since $X'|_\at=X'$ for any $X'$) and such that 
$X''\in E_s(\T)$ for all $X''\subseteq Y$ where $X'|_\at\subseteq X''|_\at$ 
(i.e., for $X'=X\subseteq X''\subseteq Y$).
Note that this are exactly the requirements for $(X,Y)$ being closed in 
$E_s(\T)$, thus for $(X,Y)\in E_u(\T)$. 
Therefore, $(X,Y)\in \Eh{\T}$ iff $(X,Y)\in E_u(\T)$, which proves the claim.
\end{proof}

Moreover, a setting where $\Ap=\An$ is termed relativized strong equivalence, 
and $\An=\emptyset$ denotes  relativized uniform equivalence. 
A 
further remark is in place, however. While we proved for uniform 
equivalence of propositional theories, that it is indifferent to whether we 
restrict additions (contexts) to sets of atoms or whether we allow for factual 
theories, $\at$-$\emptyset$-theories syntactically do not encompass factual 
theories, since negation, i.e., formulas of the form $a\IMP\bot$, are not 
permitted. One question that this raises is: would allowing factual theories 
as contexts make a difference for relativized notions of uniform 
equivalence?

The answer is by inspection of the proof of Theorem~\ref{theo:hyper} in 
connection with Proposition~\ref{prop:hyp-eval} and 
Proposition~\ref{prop:hyp-tot}. Recall that the propositions have been stated 
for extended $\Ap$-$\An$-theories. Therefore, the only-if direction of 
Theorem~\ref{theo:hyper} also holds for extended $\Ap$-$\An$-theories. Since 
the if direction just referred to $\Ap$-$\An$-theories (which, trivially,  
are extended $\Ap$-$\An$-theories too), we obtain the following.

\begin{corollary}\label{coro:hyper}
Two propositional theories  $\T_1, \T_2$ 
are \emph{relativized hyperequivalent wrt.~extended $\Ap$-$\An$-theories} 
if and only if they coincide on their \HT-hyperequivalence interpretations 
wrt.~$\Ap$ and $\An$. 
\end{corollary}

Thus, also relativized uniform equivalence is independent of whether sets of 
atoms or factual theories are permitted as contexts. More generally, for any 
notion of relativized hyperequivalence, factual theories over $\Ap$ can be 
allowed in the context without altering the notion of equivalence captured.
This holds essentially due to Proposition~\ref{prop:hyp-eval}, which 
generalizes Lemma~\ref{lemm:pearce} (Lemma~5 in~\cite{Pearce04}) in this 
respect. 

A final result establishes, that the notion of relativized hyperequivalence which 
has been introduced in this section is a proper generalization of the respective 
logic programming version to the more general case of propositional theories 
under answer-set semantics. It is a straight forward consequence of 
Theorem~\ref{theo:hyper}, since the $\Ap$-$\An$-theories in the proof of the 
if direction consist of formulas corresponding to rules with heads restricted to 
positive atoms from $\Ap$ and body atoms from $\An$.
Let us say that two propositional programs $\p_1$ and $\p_2$ are relativized 
hyperequivalent wrt.~$\Ap$ and $\An$ in the logic programming sense, in symbols
$\p_1\equivh_{\!\!lp}\ \p_2$, if and only if $\p_1\cup\p \equiva\p_2\cup\p$ for any 
program $\p$, such that $\headn{r}=\emptyset$, $\headp{r}\subseteq\Ap$, and 
$\body{r}\subseteq\An$, for all $r\in\p$.

\begin{corollary}\label{coro:relprog}
Given two programs $\p_1$ and $\p_2$, let $\Ap$ and $\An$ be sets of propositional
variables. Then, $\p_1\equivh_{\!\!lp}\ \p_2$ if and only if $\p_1\equivh\p_2$.
\end{corollary}

\section{Generalization to First-Order Theories} \label{sec:nonground}

Since the characterizations, in particular of uniform equivalence, presented in 
the previous section capture also infinite theories, they pave the way for 
generalizing this notion of equivalence to non-ground settings without any 
finiteness restrictions. In this section we study first-order theories.

As first-order theories we consider sets of sentences (closed formulas) of   
a first-order signature $\sig=\tuple{\FS, \PS}$ in the sense of classical 
first-order logic. Hence, $\FS$ and $\PS$ are pairwise disjoint sets of function 
symbols and predicate symbols with an associated arity, respectively. Elements 
of $\FS$ with arity $0$ are called object constants. A $0$-ary predicate symbol is 
a propositional constant. Formulas are constructed as usual and variable-free 
formulas or theories are called {\em ground}.
A sentence is said to be {\em factual} if it is built 
using connectives $\AND$, $\OR$, $\exists$, $\forall$, and $\neg$ 
(i.e., implications of the form $\F\IMP\bot$), only.
A theory $\T$ is factual if every sentence of $\T$ has this property.
The abbreviations introduced for propositional formulas carry over: $\F\equiv \G$ 
for $(\F\IMP \G) \AND (\G\IMP \F)$; $\neg \F$ for $\F\IMP\bot$; and $\top$ for 
$\bot\IMP\bot$.

\subsection{Static Quantified Logic of Here-and-There}

Semantically we refer to the static quantified version of here-and-there with 
decidable equality as captured axiomatically by the system 
$\sqhte$~\cite{pear-valv-06,lifs-etal-07,pear-valv-08}. It is 
characterized  by Kripke models of two worlds with a common universe 
(hence static) that interpret function symbols in the same way. 

More formally, consider a first-order interpretation $I$ of a first-order signature 
$\sig$ on a universe $\U$. We denote by $\sig^I$ the extension of $\sig$ obtained 
by adding pairwise distinct names $c_\varepsilon$ as object constants for the 
objects in the universe, 
i.e., for each $\varepsilon\in\U$. 
We write $\C_\U$ for the set $\{c_\varepsilon\mid \varepsilon\in\U\}$ and 
identify $I$ with its extension to $\sig^I$ given by 
$I(c_\varepsilon) = \varepsilon$.
Furthermore, let $t^I$ denote the value assigned  by $I$ to a ground term $t$ 
(of signature $\sig^I$), 
let $\sig_\FS$ denote the restriction of $\sig$ 
to function symbols (thus including object constants), 
and let $\BU$ be the set of atomic formulas built 
using predicates from $\PS$ and constants $\C_\U$.

We represent a first-order interpretation $I$ of $\sig$ on $\U$ 
as a pair $\tuple{I|_{\sig_\FS},I|_{\C_\U}}$,\footnote{We use angle brackets to 
distinguish from \HT-interpretations.} where $I|_{\sig_\FS}$ is the restriction 
of $I$ on function symbols, and $I|_{\C_\U}$ is the set of 
atomic formulas 
from $\BU$ which are satisfied in $I$.
Correspondingly, classical satisfaction of a sentence $\F$ by a first-order 
interpretation $\tuple{I|_{\sig_\FS},I|_{\C_\U}}$ is denoted by 
$\tuple{I|_{\sig_\FS},I|_{\C_\U}}\models \F$.
We also define a subset relation for first-order interpretations $I_1, I_2$ 
of $\sig$ on $\U$ (ie., over the same domain)
by $I_1 \subseteq I_2$ if $I_1|_{\sig_\FS}=I_2|_{\sig_\FS}$ and 
$I_1|_{\C_\U}\subseteq I_2|_{\C_\U}$.

A \QHT-interpretation of $\sig$ is a triple $\tuple{I,J,K}$, such that
($i$) $I$ is an interpretation of $\sig_\FS$ on $\U$, and 
($ii$) $J\subseteq K\subseteq\BU$.

The satisfaction of a sentence $\F$ of signature $\sig^I$ by a \QHT-interpretation 
$M=\tuple{I,J,K}$  (a \QHT-model) is defined as:
\begin{enumerate}
\item $M\models p(t_1, \ldots, t_n)$ if $p(c_{t_1^I},\ldots, c_{t_n^I})\in J$;
\item $M\models t_1 = t_2$ if $t_1^I = t_2^I$;
\item $M\not\models \bot$;
\item $M\models \F\AND \G$ if $M\models \F$ and $M\models \G$,
\item $M\models \F\OR \G$, if $M\models \F$ or $M\models \G$,
\item $M\models \F\IMP \G$ if ($i$) $M\not\models \F$ or $M\models \G$, 
and ($ii$) $\tuple{I,K}\models \F\IMP \G$\footnote{That is, $\tuple{I,K}$ satisfies $\F\IMP \G$ classically.};
\item $M\models \forall x \F(x)$ if $M\models \F(c_\varepsilon)$ and 
$\tuple{I,K}\models \F(c_\varepsilon)$ for all $\varepsilon\in\U$;
\item $M\models \exists x \F(x)$ if $M\models \F(c_\varepsilon)$ for some 
$\varepsilon\in\U$;.
\end{enumerate}

A \QHT-interpretation $M=\tuple{I,J,K}$ is 
called a {\em \QHT-countermodel} of a theory $\T$ iff $M\not\models\T$; it is 
called {\em total} if $J=K$. A total \QHT-interpretation $M=\tuple{I,K,K}$ is called a 
\emph{quantified equilibrium model $($\QEL-model$)$}  of a theory $\T$, 
iff $M\models\T$ and $M'\not\models\T$, for all QHT-interpretations 
$M'=\tuple{I,J,K}$ such that 
$J\subset K$. A first-order interpretation $\tuple{I,K}$ is an \emph{answer set} 
of $\T$ iff $M=\tuple{I,K,K}$ is a \QEL-model of a theory $\T$.

In analogy to the propositional case, we will use the following simple properties.
If $\tuple{I,J,K}\models \F$ then $\tuple{I,K,K}\models \F$; and 
$\tuple{I,J,K}\models \neg \F$ iff 
$\tuple{I,K}\models \neg \F$.

\subsection{Characterizing Equivalence by \QHT-countermodels}

We aim at generalizing uniform equivalence for first-order theories, in its most 
liberal form, which means wrt.~factual theories. For this purpose, 
we first lift Lemma~\ref{lemm:pearce}.

\begin{lemma}\label{lemm:factng}
Let $\F$ be a factual sentence. 
If $\tuple{I,J,K}\models \F$ and $J\subseteq J'\subseteq K$, then 
$\tuple{I,J',K}\models \F$.
\end{lemma}

\begin{proof}
The proof is by induction on the formula structure of $\F$. Let $M=\tuple{I,J,K}$, 
$M\models \F$, and $M'=\tuple{I,J',K}$ for some $J\subseteq J'\subseteq K$.
For the base case, consider an atomic sentence $\F$. If $\F$ is of the form 
$p(t_1, \ldots, t_n)$, then $p(c_{t_1^I},\ldots, c_{t_n^I})\in J$ because 
$M\models\F$. By the fact that $J'\supseteq J$ we conclude that 
$p(c_{t_1^I},\ldots, c_{t_n^I})\in J'$ and hence $M'\models\F$. If $\F$ is of the 
form  $t_1 = t_2$ then $M\models\F$ implies $t_1^I = t_2^I$, and thus $M'\models\F$. 
Note also that $M\models \F$ implies $\F\neq\bot$.
This proves the claim for atomic formulas.

For the induction step, assume that $M\models\F$ implies $M'\models\F$, for any 
sentence of depth $n-1$, and let $\F$ be a sentence of depth $n$. We 
show that $M\models\F$ implies $M'\models\F$. Suppose $\F$ is the conjunction or 
disjunction of two sentences $\F_1$ and $\F_2$. Then $\F_1$ and $\F_2$ are sentences 
of depth $n-1$. 
Hence, $M\models\F_1$ implies $M'\models\F_1$, and the same for $\F_2$. Therefore, if 
$M$ models both or one of the sentences then so does $M'$, which implies  
$M\models\F$ implies $M'\models\F$ if $\F$ is the conjunction or disjunction of two 
sentences. 
As for implication, since  $\F$ is factual we just need to consider the case where 
$\F$ is of the form $\F_1\IMP\bot$, i.e., $\neg\F_1$. 
Then, $M\models\neg\F_1$ iff $\tuple{I,K}\models\neg\F_1$ iff $M'\models\neg\F_1$. This 
proves $M\models\F$ implies $M'\models\F$  if $\F$ is an implication with $\bot$ as 
its consequence.
Eventually, consider a quantified sentence $\F$, 
i.e., $\F$ is of the form $\forall x \F_1(x)$ or $\exists x \F_1(x)$. 
In this case, $M\models\F$ implies $M\models\F_1(c_\varepsilon)$ and 
$\tuple{I,K}\models\F_1(c_\varepsilon)$, for all $\varepsilon\in\U$, 
respectively $M\models\F_1(c_\varepsilon)$, for some $\varepsilon\in\U$, in case of 
existential quantification. Since each of the sentences $\F_1(c_\varepsilon)$ 
is of depth $n-1$, the same is true for $M'$ by assumption, i.e., 
$M'\models\F_1(c_\varepsilon)$ and $\tuple{I,K}\models\F_1(c_\varepsilon)$, for all 
$\varepsilon\in\U$, respectively $M'\models\F_1(c_\varepsilon)$, for some 
$\varepsilon\in\U$.
It follows that  
$M\models\F$ implies $M'\models\F$ also for quantified sentences $\F$ of depth $n$, 
and therefore, for any sentence $\F$ of depth $n$. This proves the claim.
\end{proof}

The different notions of closure naturally extend to (sets of) 
\QHT-interpretations. 
In particular, 
a total \QHT-interpretation $M=\tuple{I,K,K}$ is called 
\emph{total-closed} in a set $S$ of \QHT-interpretations 
if $\tuple{I,J,K}\in S$  for every $J\subseteq K$.
A \QHT-interpretation $\tuple{I,J,K}$ is 
\emph{closed} in a set $S$ of \QHT-interpretations if $\tuple{I,J',K}\in S$ 
for every $J\subseteq J'\subseteq K$, and it is 
\emph{there-closed} in $S$ 
if $\tuple{I,K,K}\not\in S$ and $\tuple{I,J',K}\in S$ for every 
$J\subseteq J'\subset K$.

The first main result lifts the characterization of uniform equivalence for theories 
by \HT-counter\-models to the first-order case.

\begin{theorem}\label{theo:counterng}
Two 
first-order theories are uniformly equivalent iff 
they have the same sets of there-closed \QHT-countermodels.
\end{theorem}

The proof idea is the same as in the propositional case, thus for space reasons the proof 
is skipped. The same applies to 
Theorem~\ref{theo:equimodng} and 
Proposition~\ref{prop:charng} (cf.~\cite{fink-09tr} for full proofs).

We next turn to an alternative characterization by a mixture of \QHT-models and 
\QHT-counter\-models as in the propositional case. 
A \QHT-counter\-model $\tuple{I,J,K}$ of a theory $\T$ is called 
\QHT\ here-countermodel of $\T$ if $\tuple{I,K}\models \T$.
A \QHT-interpretation $\tuple{I,J,K}$ is an \QHT\ equivalence-interpretation of a 
theory $\T$, if it is a total \QHT-model of $\T$ or a \QHT\ here-countermodel 
of $\T$.
In slight abuse of notation, we reuse the notation $S_e$, $S\in\{C,E\}$ and 
$e\in\{c,a,s,u\}$, for respective sets of \QHT-interpretations, 
and arrive at the following formal result:

\begin{theorem}\label{theo:equimodng}
Two theories coincide on their \QHT-countermodels iff they have the same \QHT\ 
equivalence-interpretations, in symbols  $C_s(\T_1)=C_s(\T_2)$ iff $E_s(\T_1)=E_s(\T_2)$.
\end{theorem}

As a consequence of these two main results, we obtain an elegant, unified formal 
characterization of the different notions of equivalence for first-order theories under 
generalized answer-set semantics.

\begin{corollary}
Given two first-order theories $\T_1$ and $\T_2$, 
the following propositions are equivalent for $e\in\{c,a,s,u\}$:
$\T_1\equive\T_2$; 
$C_e(\T_1)=C_e(\T_2)$; 
$E_e(\T_1)=E_e(\T_2)$. 
\end{corollary}

Moreover, lifting the characterization of \HT-countermodels provided in 
Proposition~\ref{prop:char} to the first-order setting,
allows us to prove a property, which simplifies the treatment of extended 
signatures.

\begin{proposition}\label{prop:charng}
Let $M$ be a \QHT-interpretation over $\sig$ on $\U$. Then, 
$M\in E_s(\T)$ for a theory $\T$ iff $M\models \T_\F(M)$ for some $\F\in\T$, 
where 
$\T_\F(M)=\{\neg\neg \G\mid \G\in\T\} \cup \{\F\IMP (\neg\neg a\IMP a)\mid a\in\BU\}$.
\end{proposition}

For \QHT-models it is known that $M\models\T$ implies 
$M|_\sig\models\T$ (cf. e.g., Proposition~3 in~\cite{brui-etal-07}), hence 
$M|_\sig\not\models\T$ implies $M\not\models\T$, i.e., $M|_\sig\in C_s(\T)$ 
implies $M\in C_s(\T)$. The converse direction holds for totality preserving 
restrictions 
(the proof appeared in~\cite{fink-08} and can also be found in~\cite{fink-09tr}):

\begin{theorem}\label{theo:submodng}
Let $\T$ be a theory over $\sig$, let $\sig'\supset \sig$, and let  
$M$ a \QHT-interpretation over $\sig'$ such that $M|_\sig$ is totality 
preserving. Then,  
$M\in C_s(\T)$ implies $M|_\sig\in C_s(\T)$.
\end{theorem}

Note that this property carries over to \QHT-models, i.e., $M|_\sig\models\T$
implies $M\models\T$, if $M|_\sig$ is the restriction of $M$ to $\sig$ and 
this restriction is totality preserving. Otherwise, by the above result 
$M\not\models\T$ would imply $M|_\sig\not\models\T$. We remark that 
in~\cite{fink-08} it is erroneously stated informally that this property does 
not hold for \QHT-models, however the counter-example given there is flawed 
(Example~5 in~\cite{fink-08}).

\subsection{Relativized Hyperequivalence for First-Order Theories}

In this section we extend the notion of relativized hyperequivalence to 
first-order theories. For this purpose, we distinguish positive and negative 
occurrences of predicates in sentences. More precisely, the 
occurrence of a predicate $p$ in a sentence $\F$ is called 
\emph{positive} if $\F$ is implication free, if $p$ occurs in the 
consequent of an implication in $\F$, or if $\F$ is of the form 
$(\F_1\IMP\F_2)\IMP\F_3$ and $p$ occurs in $\F_1$. An occurrence of $p$ is called 
\emph{negative} if $p$ occurs in the antecedent of an implication. 
The notion of positive and negative occurrence is again extended to 
(sub-)sentences in the obvious way.

Let $\T$ be a first-order theory over $\sig=\tuple{\FS, \Lp\cup\Ln}$, where 
$\Lp$ and $\Ln$ are sets of predicate symbols with an associated arity, such 
that if a predicate symbol $p$ occurs in both $\Lp$ and $\Ln$, then it is also 
associated the same arity. We say that $\T$ is an \emph{$\Lp$-$\Ln$-theory} if 
its sentences have positive occurrences of predicates from $\Lp$, and negative 
occurrences of predicates from $\Ln$, only. 
As in the propositional case, $\bot$ is allowed 
to appear positively and negatively, and the same holds for equality in the 
first-order case. Moreover, an $\Lp$-$\Ln$-theory is called 
\emph{extended}, if additionally factual formulas over $\Lp$ 
are permitted. 

\begin{definition}\label{def:fo-hyperequiv}
Two first-order theories  $\T_1, \T_2$ over $\sig$ are called
\emph{relativized hyperequivalent wrt.~$\Lp$ and $\Ln$}, symbolically 
$\T_1\equivhl\T_2$, iff for any $\Lp$-$\Ln$-theory $\T$ over 
$\sig'\supseteq\sig$, $\T_1\cup \T$ and  $\T_2\cup \T$ are answer-set 
equivalent. 
\end{definition}

The properties proven for \HT-interpretations and extended $\Ap$-$\An$-theories
in the propositional case, carry over to \QHT-interpretations and extended 
$\Lp$-$\Ln$-theories in a straight forward manner.

\begin{proposition}\label{prop:fo-hyp-eval}
Consider an extended first-order $\Lp$-$\Ln$-theory $\T$, 
and a \QHT-interpretation $\tuple{I,J,K}$. Then, 
$\tuple{I,J,K}\models \T$ implies 
$\tuple{I,J',K}\models \T$, 
for all $J'\subseteq K$ such that 
$J|_{\Lp} \subseteq J'|_{\Lp}$ and 
$J'|_{\Ln} \subseteq J|_{\Ln}$. 
\end{proposition}

The proof is lengthy and does not convey particular new insights, therefore 
it is skipped here (cf.~\cite{fink-09tr}). The main differences to the 
propositional case concern the treatment of equality of terms and that 
quantification has to be taken into account. The former depends solely on 
the interpretation part $I$, which is the same for the \QHT-interpretations 
under consideration, and thus has no further influence on the argument. The 
latter, is a further case to be considered in the inductive argument, however 
one that reduces easily to the respective induction hypotheses. The remainder 
simply mirrors the propositional case, with the polarity being considered on 
the predicate level, rather than for propositional variables. The same holds 
for the proofs of the 
remaining results in this section.

\begin{proposition}\label{prop:fo-hyp-tot}
Consider an extended first-order $\Lp$-$\Ln$-theory $\T$, 
and a total \QHT-interpretation $\tuple{I,K,K}$. Then, 
$\tuple{I,K,K}\models \T$ implies $\tuple{I,J',K}\models \T$, 
for all $J'\subseteq K$ such that 
$J'|_{\Lp} = K|_{\Lp}$. 
\end{proposition}

Having lifted the essential properties to the case of  $\Lp$-$\Ln$-theories, 
it comes at no surprise that we end up with respective closure conditions 
for \QHT-equivalence interpretations.

\begin{definition}\label{def:fo-hyperclosure}
Given a first-order theory $\T$ over $\sig$, sets of predicate symbols  
$\Lp\subseteq\sig'$, $\Ln\subseteq\sig'$, $\sig'\supseteq\sig$, and 
a \QHT-interpretation $M=\tuple{I,J,K}$, 
we say that
\begin{itemize}
\item $\tuple{I,K,K}$ is \emph{$\Lp$-total} iff $\tuple{I,K|_{\Lp},K}$ is 
closed in $E_s(\T)$;
\item 
  $M$ is \emph{$\Lp$-closed} in $E_s(\T)$ iff 
   $\tuple{I,J',K}\in E_s(\T)$, for all $J'\subseteq K$ such that
       $J|_{\Lp}\subseteq J'|_{\Lp}$ and $J'|_{\Ln} \subseteq J|_{\Ln}$. 
\end{itemize}
\end{definition}

Also the characteristic structures for a semantic characterization are 
defined in straight-forward analogy.

\begin{definition}\label{def:fo-hypercmodel}
A \QHT-interpretation $M=\tuple{I,J,K}$ is a \emph{\QHT-hyperequivalence 
interpretation wrt.~$\Lp$ and $\Ln$} of a first-order theory $\T$
iff $\tuple{I,K,K}$ is $\Lp$-total and 
there exists a \QHT-inter\-pretation $\tuple{I,J',K}$ such that 
$J=J'|_{\Lp\cup\Ln}$ and $\tuple{I,J',K}$ is $\Lp$-closed in $E_s(\T)$.

\noindent
The set of \QHT-hyperequivalence interpretations wrt.~$\Lp$ and $\Ln$ 
of a first-order theory $\T$ is denoted by $\Ehl{\T}$.
\end{definition}

Eventually, we arrive at a characterization of relativized hyperequivalence 
for general first-order theories under answer-set semantics, where contexts 
are restricted on the predicate level.

\begin{theorem}\label{theo:fo-hyper}
Two first-order theories  $\T_1, \T_2$ 
are relativized hyperequivalent wrt.~$\Lp$ and $\Ln$ if and only if 
they coincide on their \QHT-hyperequivalence interpretations 
wrt.~$\Lp$ and $\Ln$, 
symbolically  $\T_1\equivhl\T_2$ iff $\Ehl{\T_1}=\Ehl{\T_2}$.
\end{theorem}

In the same way as for propositional theories, the prominent notions of 
equivalence are obtained as special cases, and the framework gives rise to 
relativized notions of strong and uniform equivalence for general first-order 
theories under answer-set semantics. Also in analogy, the role of factual 
theories is governed by Proposition~\ref{prop:fo-hyp-eval}, yielding the 
following:

\begin{corollary}\label{coro:fo-hyper}
Two first-order theories  $\T_1, \T_2$ 
are \emph{relativized hyperequivalent wrt.~extended $\Lp$-$\Ln$-theo\-ries} 
if and only if they coincide on their \QHT-hyperequivalence interpretations 
wrt.~$\Lp$ and $\Ln$. 
\end{corollary}

\section{Non-ground Logic Programs} \label{sec:lp}

In this section we apply the characterizations obtained for first-order theories 
to non-ground logic programs under various extended semantics---compared to the 
traditional semantics in terms of Herbrand interpretations. 
For a proper treatment of these issues, further 
background is 
required and introduced (succinctly, but at sufficient detail) below.

In non-ground logic programming, we restrict to a function-free first-order 
signature $\sig=\tuple{\FS,\PS}$ (i.e., $\FS$ contains object constants only) 
without equality. 
A \emph{program} $\p$ (over $\sig$) is a set of rules (over $\sig$) of the 
form (\ref{form:rule}). 
A rule $r$ is \emph{safe} if each variable 
occurring in $\head{r}\cup\bodyn{r}$ also occurs in $\bodyp{r}$;
a rule $r$ is \emph{ground}, if all atoms occurring in it are ground.
A program is safe, respectively ground, if all of its rules enjoy this property.

Given $\p$ over $\sig$ and a universe $\U$, let $\sig^\U$ be the extension of 
$\sig$ as before. The 
\emph{grounding of $\p$} wrt.~$\U$ and an interpretation $I|_{\sig_\FS}$ of 
$\sig_\FS$ on $\U$ is defined as the set $\gr_\U(\p,I|_{\sig_\FS})$ of ground rules 
obtained from $r\in\p$ by ($i$) replacing any constant $c$ in $r$ by 
$c_\varepsilon$ such that $I|_{\sig_\FS}(c)=\varepsilon$, and 
($ii$) all possible substitutions of elements in $\C_\U$ for the variables in $r$.

Adapted from~\cite{gelf-lifs-91}, 
the \emph{reduct} of a program $\p$ with respect to a first-order interpretation 
$I=\tuple{I|_{\sig_\FS},I|_{\C_\U}}$ on universe $\U$, in symbols  
$\gr_\U(\p,I|_{\sig_\FS})^I$, is given by the set of rules 
\vspace*{-.5ex}
$$a_1\vee\cdots\vee a_k \ \la\ b_1,\ldots, b_m, $$
obtained from rules in $\gr_\U(\p,I|_{\sig_\FS})$ of the form (\ref{form:rule}), 
such that $I\models a_i$ for all 
$k< i\leq l$ and $I\not\models b_j$ for all $m< j\leq n$.

A first-order interpretation $I$ satisfies a rule $r$, $I\models r$, iff $I\models \T_r$, where 
$\T_r = \forall \vec{x} (\beta_r \IMP \alpha_r)$, $\vec{x}$ are the free 
variables in $r$, $\alpha_r$ is the disjunction of $\head{r}$, and $\beta_r$ is the 
conjunction of $\body{r}$. It satisfies a program $\p$, symbolically $I\models\p$, 
iff it satisfies every $r\in\p$, i.e., if $I\models \T_\p$, where 
$\T_\p=\bigcup_{r\in\p} \T_r$.

A first-order interpretation $I$ is called a \emph{generalized answer set} of $\p$ 
iff 
it satisfies $\gr_\U(\p,I|_{\sig_\FS})^I$ 
and it is subset minimal among the 
interpretations of $\sig$ on $\U$ with this property.

Traditionally, only \emph{Herbrand interpretations} are considered as the answer 
sets of a logic program. The set of all (object) constants occurring in $\p$ is 
called the \emph{Herbrand universe} of $\p$, symbolically $\HU$.
If no constant appears in $\p$, then $\HU=\{c\}$, for an arbitrary constant $c$. 
A Herbrand interpretation is any interpretation $I$ of $\sig_\HU=\tuple{\HU,\PS}$ 
on $\HU$ interpreting object constants by identity, $\mathit{id}$, i.e., 
$I(c)=\mathit{id}(c)=c$ for all $c\in\HU$.
A Herbrand interpretation $I$ is an \emph{ordinary answer set} of $\p$ iff it is 
subset minimal among the interpretations of $\sig_\HU$ on $\HU$ satisfying 
$\gr_\HU(\p,\mathit{id})^I$.

Furthermore, an \emph{extended Herbrand interpretation} is an interpretation of $\sig$
on $\U\supseteq\FS$ interpreting object constants by identity. 
An extended Herbrand interpretation $I$ is an 
\emph{open answer set}~\cite{heym-etal-07} of $\p$ iff 
it is subset minimal among the interpretations of $\sig$ on $\U$ satisfying 
$\gr_\U(\p,\mathit{id})^I$.

Note that since we consider programs without equality, 
we semantically resort to the logic \sqht, which results from \sqhte\ by dropping 
the axioms for equality. Concerning Kripke models, however, in slight abuse of 
notation, we reuse 
\QHT-models as defined for the general case.
A \QHT-interpretation $M=\tuple{I,J,K}$ is called an (extended) \QHT\  
Herbrand interpretation, if $\tuple{I,K}$ is an (extended) Herbrand interpretation.
Given a program $\p$, $\tuple{I,K}$ is a generalized answer set of $\p$ iff 
$\tuple{I,K,K}$ is a \QEL-model of $\T_\p$, and $\tuple{I,K}$ is an open, respectively 
ordinary, answer set of $\p$ iff $\tuple{I,K,K}$ is an extended Herbrand, 
respectively Herbrand, \QEL-model of $\T_\p$. Notice that the static interpretation 
of constants introduced by Item~($i$) of the grounding process is essential for 
this correspondences in terms of \sqht.
%
Abusing 
notation, we further on identify $\p$ and  $\T_\p$.

As already mentioned for propositional programs, uniform equivalence 
is usually 
understood wrt.~sets of \emph{ground facts} (i.e., ground atoms). Obviously, 
uniform equivalence wrt.~factual theories implies uniform equivalence wrt.~
ground atoms. We show the converse direction (lifting 
Theorem~2 in~\cite{Pearce04}, for a proof see ~\cite{fink-09tr}).

\begin{proposition}\label{prop:heads}
Given two programs $\p_1,\p_2$, then $\p_1\equivu\p_2$ iff 
$(\p_1\cup A)\equiva (\p_2\cup A)$, for any set of ground atoms $A$.
\end{proposition}

Thus, 
there is no difference whether we consider uniform equivalence 
wrt.~sets of 
ground facts or factual theories. 
Since one 
can also consider sets of clauses, i.e. disjunctions of atomic formulas 
and their negations, which is a more suitable representation of facts 
according to the definition of program rules in this article,  
we adopt the following terminology.
A rule $r$ is called a \emph{fact} if $\body{r}=\emptyset$, and 
a \emph{factual program} is a set of 
facts.  Then, by our result $\p_1\equivu\p_2$ holds for programs $\p_1,\p_2$ iff 
$(\p_1\cup \p)\equiva (\p_2\cup \p)$, for any factual program $\p$.

\subsection{Uniform Equivalence under Herbrand Interpretations}

The results in the previous section generalize the notion of 
uniform equivalence to programs under generalized open answer-set semantics and 
provide alternative characterizations  for other notions of equivalence. 
They apply to programs under open answer-set semantics and ordinary 
answer-set semantics, when \QHT-interpretations are restricted to extended Herbrand 
interpretations and Herbrand interpretations, respectively. In order to capture 
strong and uniform equivalence under ordinary answer-set semantics correctly, 
interpretations under the Standard Name Assumption (SNA) have to be considered, 
accounting for the potential extensions. 
For programs 
$\p_1$ and $\p_2$ and $e\in\{c,a,s,u\}$, we use  $\p_1\equive^\HE\p_2$ and  
$\p_1\equive^\HU\p_2$ to denote (classical, answer-set, strong, or uniform) equivalence 
under open answer-set semantics and ordinary answer-set semantics, respectively.

\begin{corollary}
Given two programs $\p_1$ and $\p_2$, it holds that 
\vspace*{-.5ex}
\begin{itemize}
\item
$\p_1\equive^\HE\p_2$, $C_e^\HE(\p_1)=C_e^\HE(\p_2)$, and $E_e^\HE(\p_1)=E_e^\HE(\p_2)$ 
are equivalent; and 
\item
$\p_1\equive^\HU\p_2$, $C_e^\HU(\p_1)=C_e^\HU(\p_2)$, and $E_e^\HU(\p_1)=E_e^\HU(\p_2)$ 
are equivalent; 
\end{itemize}
\vspace*{-.5ex}
where 
$e\in\{c,a,s,u\}$, superscript $\HE$ denotes the 
restriction to extended Herbrand interpretations, and 
superscript $\HU$ denotes the 
restriction to Herbrand interpretations for $e\in\{c,a\}$, 
respectively to SNA interpretations for $e\in\{s,u\}$.
\end{corollary}

For safe programs 
open answer sets and 
ordinary answer sets 
coincide~\cite{brui-etal-07}. Note that a fact is safe if it is ground. 
We obtain that uniform equivalence coincides under the two semantics even for 
programs that are not safe. 
Intuitively, the potential addition of arbitrary facts accounts 
for the difference in the semantics since it requires to consider larger domains 
than the Herbrand universe.\footnote{Note that this 
also holds for 
\sqhte\ with functions and the result could be strengthened accordingly.} 

\begin{theorem}\label{theo:openord}
Let $\p_1, \p_2$ be programs over $\sig$. Then, 
$\p_1\equivu^\HE\p_2$ iff $\p_1\equivu^\HU\p_2$.
\end{theorem}

\begin{proof}
The only-if direction is trivial. For the if direction, towards a contradiction 
assume that $\p_1\equivu^\HU\p_2$ and $\p_1\not\equivu^\HE\p_2$. Let $\p$ be a 
factual program such that $M=\tuple{\mathit{id},K,K}$ is 
an extended Herbrand \QHT-interpretation over $\sig'\supseteq\sig$ on $\U'$, such
that $M$ is in $E_a^\HE(\p_1\cup \p)$, but $M\not\in E_a^\HE(\p_2\cup \p)$. 
Consider the signature $\sig_{\U'}=\tuple{\U',\sig'_\PS\cup\{d\}}$, where $\sig'_\PS$ 
are the predicate symbols of $\sig'$, and $d\not\in\sig'_\PS$ is a fresh unary 
predicate symbol. Clearly,  $\sig_{\U'}\supset\sig'$. Furthermore let 
$\p'_1=\p_1\cup\p\cup\{d(X)\}$, $\p'_2=\p_2\cup\p\cup\{d(X)\}$, and 
$K'=K\cup \{d(c)\mid c\in\U'\}$.
We show that $M'=\tuple{\mathit{id},K',K'}$ is 
in $E_a^\HU(\p'_1)$, but $M'\not\in E_a^\HU(\p'_2)$.
Since $M\models \p_1\cup\p$ and no sentence in $\p_1\cup\p$ involves
$d$, we conclude $M'\models \p_1\cup\p$. By construction, $M'$ is also a \QHT-model 
of $d(X)$, hence $M'\models \p'_1$. Moreover, 
$\tuple{\mathit{id},J,K}\not\models\p_1\cup\p$, for every $J\subset K$. Therefore, 
for every $J'=J\cup \{d(c)\mid c\in\U'\}$ such that  $J\subset K$, 
$\tuple{\mathit{id},J',K'}\not\models\p'_1$. So let us consider proper subsets $J'$
of $K'$ such that $K\subseteq J$, i.e.,  $J'\subset\{d(c)\mid c\in\U'\}$. 
In this case $\tuple{\mathit{id},J',K'}\not\models d(X)$, and again 
$\tuple{\mathit{id},J',K'}\not\models\p'_1$. This proves that $M'$ is 
in $E_a^\HU(\p'_1)$.
On the other hand, if $M\not\models\p_2\cup\p$, then  
$M\not\models\p_2$, and since 
no sentence in $\p_2$ involves $d$, we conclude $M'\not\models \p_2$, thus 
$M'\not\models \p'_2$. 
If  $M\models\p_2\cup\p$, then $\tuple{\mathit{id},J,K}\models\p_2\cup\p$ for 
some $J\subset K$. 
Consider $J'=J\cup \{d(c)\mid c\in\U'\}$. Since $J\subset K$, it holds that
$J'\subset K'$, and since no sentence in $\p_2\cup\p$ involves $d$,
$\tuple{\mathit{id},J',K'}\models\p_2\cup\p$. Moroever, 
$\tuple{\mathit{id},J',K'}\models\{d(X)\}$ by construction, hence 
$\tuple{\mathit{id},J',K'}\models\p'_2$.
This proves $M'\not\in E_a^\HU(\p'_2)$.
Note that $\p\cup\{d(X)\}$ is a factual program; contradiction. 
\end{proof}

Finally, we turn to the practically relevant setting of finite, 
possibly unsafe,
programs under Herbrand interpretations, i.e., ordinary (and open) 
answer-set semantics. 
For finite programs, uniform equivalence can be characterized 
by \HT-models of the grounding, also 
for infinite domains.
In other words, the problems of ``infinite chains'' as in Example~\ref{ex:inf} 
cannot be generated by the process of grounding. 
Note that the restriction to finite programs also applies to the programs considered to be potentially added.

\begin{theorem}\label{theo:UE}
Let $\p_1, \p_2$ be finite 
programs over $\sig$. Then, 
$\p_1\equivu^\HU\p_2$ 
iff $\p_1$ and $\p_2$ have the same ($i$) total and ($ii$) maximal, non-total  
extended Herbrand \QHT-models. 
\end{theorem}

\begin{proof}
The only-if direction is obvious. If $\p_1\equivu^\HU\p_2$ then also 
$\p_1\equivu^\HE\p_2$ by Theorem~\ref{theo:openord}. 
This means that $\p_1$ and $\p_2$ 
have ($i$) the same total extended Herbrand \QHT-models, as well as the same 
sets of closed extended Herbrand \QHT\ equivalence interpretations, 
and thus ($ii$) the same maximal, non-total extended Herbrand \QHT-models.

For the if direction, assume that $\p_1$ and $\p_2$ have the same total and 
the same maximal, non-total extended Herbrand \QHT-models but, towards a 
contradiction, that $\p_1\not\equivu^\HU\p_2$. Then, there exists a finite 
factual program $\p$, such that $(\p_1\cup\p)\not\equiva^\HU(\p_2\cup\p)$. 
W.l.o.g.~let $M=\tuple{I,K,K}$ over $\sig'\supseteq\sig$ be 
in $E_a^\HU(\p_1\cup\p)$ and $M\not\in E_a^\HU(\p_2\cup\p)$.
Let $\HU$ denote the Herbrand universe of $\p_1\cup\p$. 
Since $\p_1$ and $\p$ are finite, 
$\HU$ is finite and so is $\gr_\HU(\p_1\cup\p,\mathit{id})$.
Therefore, by minimality, $K$ is finite as well. 
Note also, that  $M$ is a total extended Herbrand QHT-model of $\p_1$. By 
hypothesis ($i$), $\p_1$ and $\p_2$ have the same total extended Herbrand 
\QHT-models. Thus, $M$ is also a total extended Herbrand QHT-model of $\p_2$.
Moreover, there exists a \QHT-interpretation 
$M'=\tuple{I,J,K}$, such that $J\subset K$ and  $M'\models (\p_2\cup\p)$, 
hence $M'\models\p_2$. Since $K$ is finite, we conclude that $\p_2$ has a 
maximal, non-total \QHT-model $M''=\tuple{I,J'',K}$, such that 
$J'\subseteq J''\subset K$.
We show that this is not the case for $\p_1$.
$M'\models (\p_2\cup\p)$ implies $M'\models\p$. Since $\p$ is a factual program,
by Lemma~\ref{lemm:factng} we conclude that $M''\models\p$. 
However $M''\not\models\p_1\cup\p$, because 
$M\in E_a^\HU(\p_1\cup\p)$.
Taken together, $M''\models\p$ and $M''\not\models\p_1\cup\p$ implies 
$M''\not\models\p_1$. Therefore, $M''$ is not a maximal, non-total \QHT-model of
$\p_1$. Observing that $M''$ is an Herbrand \QHT-model over $\sig'$ and 
$\sig'\supseteq\sig$, we conclude that $M''$ is a maximal non-total extended 
Herbrand \QHT-model of $\p_2$, but not of $\p_1$. Contradiction.
\end{proof}

\section{Conclusion} \label{sec:conclusion}

Countermodels in equilibrium logic have recently been used 
by~\citeN{caba-ferr-07} 
to show that propositional disjunctive logic programs with negation in the head 
are strongly equivalent to propositional theories, and 
by~\citeN{caba-etal-07} to generate a minimal logic 
program for a given propositional theory.

By means of QEL, 
in~\cite{lifs-etal-07}, 
the notion of strong equivalence has been extended to 
first-order theories with equality, under the generalized notion of answer set we 
have adopted. 
\QEL\ has also been shown to capture open answer-sets~\cite{heym-etal-07} and 
generalized open answer-sets~\cite{heym-etal-08}, 
and 
is a promising framework for 
hybrid knowledge 
bases, providing a unified semantics  encompassing classical logic as 
well as disjunctive logic programs under the answer-set 
semantics~\cite{brui-etal-07}.

Our results extend these foundations for the research of semantic properties 
in these generalized settings. First, they complete the picture 
concerning the prominent notions of equivalence by making uniform equivalence 
amenable to these generalized settings without any 
finiteness restrictions, in particular on the domain. In addition, 
the developed notion of relativized hyperequivalence interpretation  
provides a means for the study of more specific semantic relationships under 
generalized answer-set semantics.
Thus, a general and uniform model-theoretic framework is achieved for the 
characterization of various notions of equivalence studied in ASP.
We have also shown that for finite 
programs, i.e., those programs solvers are able to deal with, 
infinite domains do not cause the problems observed for infinite propositional 
programs, when dealing with uniform equivalence in terms of \HT-models of the 
grounding.

An intersting theoretical problem for further work 
is to 
consider equivalences and correspondence under projections of answer 
sets~\cite{eite-etal-05,oets-etal-07,pueh-etal-08,pueh-tomp-09}. 
It is not difficult to apply existing techniques to 
our characterizations in order to obtain characterizations for projective 
versions of uniform and strong equivalence, and 
for relativized 
notions thereof, i.e., as long as the same alphabet is permitted for positive 
and negative occurrences in the context.
However, it is not trivial to characterize projective versions of relativized 
hyperequivalence in the general case, something which also has not been 
considered for propositional logic programs so far.

Concerning the application of our results, there is ongoing work 
on combining ontologies and nonmonotonic rules, 
an important issue in knowledge representation and reasoning for the Semantic 
Web. The study of equivalences and correspondences under an appropriate 
(unifying) semantics, such as the generalizations of answer-set semantics 
characterized by \QEL, constitute a highly relevant topic for research in 
this application domain~\cite{fink-pear-09}. Like for Datalog, uniform 
equivalence may serve investigations on query equivalence and query containment 
in these hybrid settings, and due to the combination of two formalisms, more 
specific notions of equivalence are needed to obtain the intended notions of 
correspondence. 
While our characterizations serve as a basis for these investigations, in 
particular the simplified treatment of extended signatures for (equivalence) 
interpretations is expected to be of avail, when considering separate 
alphabets.

On the foundational level, 
our results raise 
the interesting question whether 
extensions of 
intuitionistic logics  
that allow for a direct characterization of countermodels, or equivalence 
interpretations, would provide 
a more suitable formal apparatus for the 
study of (at least notions of uniform) equivalences in ASP.


\subsubsection*{Acknowledgements}

I am grateful 
to the anonymous reviewers for suggestions 
to improve this article. 
This work was 
supported by the Austrian Science Fund (FWF) 
under grants 
P18019 and P20841, and by the Vienna Science and Technology Fund (WWTF) 
under project ICT08-020. 


\ifmakebbl 

\bibliographystyle{acmtrans}
\bibliography{corr.bib}

\else

\fi

\end{document}